\documentclass[twoside,11pt]{article}

\usepackage{graphicx}
\usepackage[utf8]{inputenc} % allow utf-8 input
\usepackage[T1]{fontenc}    % use 8-bit T1 fonts
\usepackage{url}            % simple URL typesetting
\usepackage{booktabs}       % professional-quality tables
\usepackage{amsfonts}       % blackboard math symbols
\usepackage{nicefrac}       % compact symbols for 1/2, etc.
\usepackage{geometry}
\usepackage{caption}
\usepackage{subcaption}

\usepackage[toc,page]{appendix}

\usepackage{jmlr2e}
\usepackage{amsmath}
\usepackage{bbding}
\usepackage{enumitem}
\usepackage{mathtools}
\DeclareMathOperator*{\argmin}{arg\,min}

\newcommand{\R}{\ensuremath{\mathbb{R}}}
\newcommand{\Rd}{\ensuremath{\mathbb{R}^d}}
\newcommand{\N}{\ensuremath{\mathbb{N}}}
\newcommand{\Circ}{\ensuremath{\mathbb{S}}}

\newcommand{\pmodel}{\ensuremath{\mathcal{P}_{\X, \Y, \Theta, \phi}}}
\renewcommand{\O}{\ensuremath{\mathcal{O}}}
\newcommand{\loss}{\ensuremath{\mathcal{L}}}
\newcommand{\leaf}{\ensuremath{{L}}}
\newcommand{\atlas}{\ensuremath{\mathcal{A}}}
\newcommand{\A}{\ensuremath{\mathcal{A}}}

\newcommand{\F}{\ensuremath{\mathcal{F}}}
\newcommand{\X}{\ensuremath{\mathcal{X}}}
\newcommand{\Y}{\ensuremath{\mathcal{Y}}}
\newcommand{\Top}{\ensuremath{\mathcal{O}}}
\newcommand{\T}{\ensuremath{\mathcal{T}}}
\newcommand{\ball}{\ensuremath{\mathcal{B}}}
\newcommand{\model}{\ensuremath{\mathcal{M}}}
\newcommand{\G}{\ensuremath{\mathcal{G}}}
\newcommand{\task}{\ensuremath{\mathcal{T}}}
\newcommand{\structure}{\ensuremath{\mathfrak{S}}}
\newcommand{\dataset}{\ensuremath{\mathfrak{D}}}
\newcommand{\process}{\ensuremath{\mathfrak{P}}}
\newcommand{\group}{\ensuremath{\mathcal{G}}}
\newcommand{\sequence}{\ensuremath{\mathcal{S}}}
\newcommand{\orbit}{\ensuremath{\mathcal{O}}}
\newcommand{\foliation}{\ensuremath{\mathcal{F}}}
\newcommand{\learning}{\ensuremath{\mathfrak{L}}}
\newcommand{\learningtransfer}{\ensuremath{\mathfrak{LT}}}

\newcommand{\partition}{\ensuremath{\mathcal{P}}}
\renewcommand{\index}{\ensuremath{\mathcal{I}}}
\renewcommand{\|}{\ensuremath{\:\big|\:}}

\jmlrheading{1}{2021}{1-48}{4/00}{10/00}{meila00a}{Janith Petangoda, Marc P. Deisenroth, Nicholas A. M. Monk}

\ShortHeadings{Learning to Transfer: A Foliated Theory}{Petangoda, Deisenroth, and Monk}

\firstpageno{1}

\begin{document}
\title{Learning to Transfer: A Foliated Theory}
% \title{Understanding the Learning of Generalisable Latent Processes Through the Lens of Differential Geometry}

\author{\name Janith Petangoda \email jcp17@imperial.ac.uk \\
       \addr Department of Computing\\
	   Imperial College London\\
	   London, United Kingdom
	   \AND
	   \name Marc Peter Deisenroth \email m.deisenroth@ucl.ac.uk \\
       \addr Centre for Artificial Intelligence\\
	   University College London\\
	   London, United Kingdom
       \AND
       \name Nicholas A. M. Monk \email n.monk@sheffield.ac.uk \\
       \addr School of Mathematics and Statisitics\\
	   University of Sheffield\\
	   Sheffield, United Kingdom
	   }
\editor{To Be Added}

\maketitle

\begin{abstract}%   <- trailing '%' for backward compatibility of .sty file
	Learning to transfer considers learning solutions to tasks in a such way that relevant knowledge can be transferred from known task solutions to new, \emph{related} tasks. This is important for general learning, as well as for improving the efficiency of the learning process. While techniques for learning to transfer have been studied experimentally, we still lack a foundational description of the problem that exposes what related tasks are, and how relationships between tasks can be exploited constructively. In this work, we introduce a framework using the differential geometric theory of foliations that provides such a foundation.
\end{abstract}


\section{Introduction} \label{sec:introduction}
In human learning, transfer is key to the efficiency in learning that we exhibit \citep{Ellis1965,Schunk2012}. Transfer allows us to apply knowledge from past experiences when learning solutions to new problems, and thereby improve the efficiency of this learning; this applies to both physical and intellectual activities. We know, for example, that knowing how to walk \emph{informs} us about how to run, and that abstract mathematical thinking can apply to many, varied fields. Learning to transfer in machine learning (ML)  is the application of such transfer in the types of learning problems we apply ML methods to. 

Learning to transfer in ML is not a new concept; initial discussions have appeared perhaps since the 1980s and 1990s in work such as \cite{Mitchell1980, Schmidhuber1995}. The inspiration for learning using transfer almost certainly stemmed from noticing transfer in human learning \citep{Ellis1965, Schunk2012}. Transfer can be thought of as a form of generalisation; \cite{Mitchell1980} describes the need for biases in generalisation in single task problems. The type of transfer that we consider is generalisation between tasks.  Work in the field progressed through suggestions for lifelong learning \citep{Thrun1994, Thrun1995}, multi-task learning \citep{Caruana1997, Silver1996}, and learning internal representations of biases, and inductive bias transfer \citep{Baxter1995, Baxter2000}. An idea of meta-learning, or learning to learn, was written by \cite{Schmidhuber1995,Schmidhuber1996}. More modern approaches that have built on these works include advances in deep learning based meta-learning \citep{Vilata2002, Hospedales2020}, few-shot learning \citep{Wang2020}, deep multitask learning \citep{Ruder2017} and neural architecture search \citep{Elsken2018}. In these, transfer allows us to learn using less data, computational resources, and time by leveraging relevant information. In the case of multitask learning, it can also lead to better generalisation \citep{Caruana1997}.

\begin{figure}[ht]
    \centering
    \includegraphics[width=0.6\textwidth]{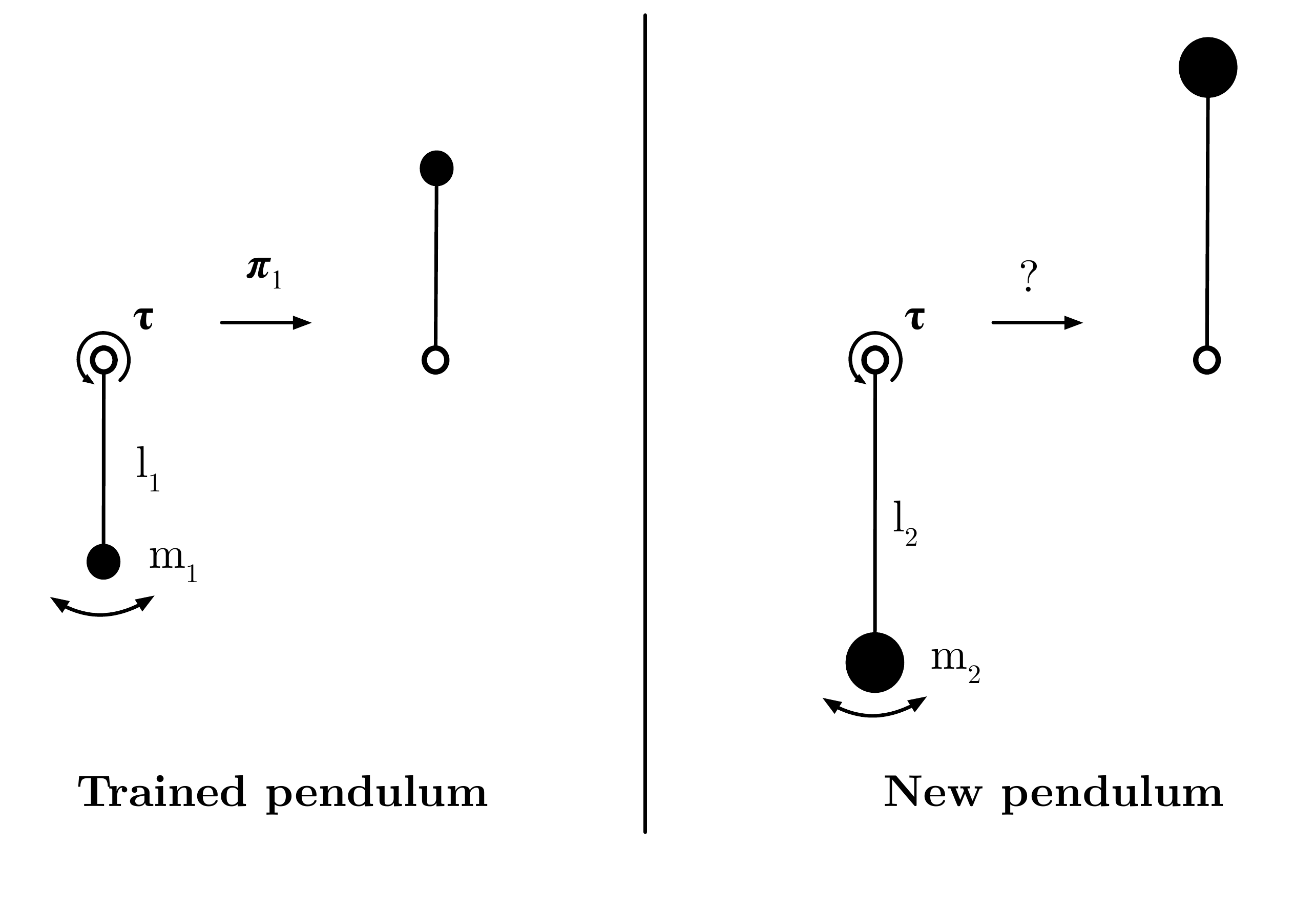}
    
    \caption{An example of a transfer problem. Having trained a policy $\pi_1$, which can swing up a pendulum with mass $m_1$ and length $l_1$ (left pane), we want to learn a policy $\pi_2$ for a new pendulum with properties $m_2$ and $l_2$ (right pane).}
     \label{fig:pendulum_example}
\end{figure}

An exemplary ML problem where transfer can be applied to is shown in Figure \ref{fig:pendulum_example}. Suppose we have learned how to swing a simple pendulum $p_1$ from its bottom-most position to its top-most position using clamped torques applied at the pivot; see Figure \ref{fig:pendulum_example}. This control strategy is denoted $\pi_1$. The defining characteristics of this particular pendulum are: 
\begin{itemize}
    \item it is a pendulum,
    \item it has a rod of length $l_1$, and 
    \item it has a point mass of $m_1$ at the end of this rod.   
\end{itemize}
Further suppose that we are given a new pendulum $p_2$ with a rod of length $l_2 \neq l_1$, and a point mass of $m_2 \neq m_1$. That is, when compared to the first pendulum, $p_2$ has the same characteristics listed previously, and is \emph{only} different in length and mass. Thus, if we wanted to learn a control policy $\pi_2$ that can swing up $p_2$, it is intuitive to assume that there is \emph{something} about the learned policy of the first, related pendulum $p_1$ that can be used to make finding such a policy easier than learning from scratch. 

%Reiterating the transfer problem
This is \emph{transfer}, and \emph{learning to transfer} is to learn $\pi_1$ in such a way that learning $\pi_2$ can exploit the \emph{relatedness} of $p_1$ and $p_2$. To reiterate, we are given tasks that are, in some sense of the word, \emph{related} to each other. Thus, we expect that the solutions to such tasks are also \emph{related} to each other in a similar sense. We would then like to exploit this relationship when learning a solution to one problem once a solution to the other, related problem is known. We expect that this exploitation will make the learning of the second solution easier than if we were to attempt learning without it. Learning to transfer, therefore, is the process of identifying such a relationship between tasks, and applying it to learning related solutions.

%Current work lacks understanding
There are many methods in the literature that can solve such problems \citep{Finn2017,Steindor2018,Janith2019}. Specific works, e.g., by \citet{Finn2017} and \citet{Steindor2018}, have been shown to be very powerful. However, most modern works are focused on obtaining working models and algorithms, rather than developing an understanding of the necessary structural and elementary components of learning to transfer. 
While the works mentioned above have shown wonderful and useful properties, it is not clear \emph{why} they work the way they do, nor even a good way of quantifying in \emph{what ways} they do work well. Having a concrete description of how transfer can be achieved allows us to measure how well it can be done; for example, how can we measure how easily transfer is possible? The practical consequences of lacking such an understanding is that we cannot \emph{deliberately} adjust our techniques and approaches for application in novel and/or more complex problems. 

This is where this article comes into play. This article provides a unifying framework for approaching problems of learning to transfer. In particular, we pose the transfer problem in terms of a mathematical framework that describes the structural building blocks of learning to transfer. 

As discussed, transfer involves leveraging a \emph{known relationship} between tasks to find solutions that are similarly related. We will isolate the relevant properties of such a relationship, and characterise them mathematically. In particular, we will see that this naturally leads us to the differential geometric notion of a foliation. Following this, we will show that some well known classes of models for transfer learning implicitly make use of this structure. 

The key goal of this article is to introduce this foliated theory of learning to transfer as a useful point of view of the problem. The framework aims to provide a conceptual guideline for thinking about problems of transfer, and to allow researchers and thinkers in the field to answer questions that include: 
\begin{enumerate}
    \item What does it mean for tasks or problems to be \emph{related}?
    \item  What does it mean to transfer between related tasks?
    \item How can the relationship between tasks be represented? 
    \item Can we learn this relationship? How similar are they?
    % \item How can we principally measure the ease of transfer, or transferrabilty? That is, how much easier does know the relationship between tasks make the learning of related or similar tasks?
\end{enumerate}
We will begin by describing how the familiar learning of a single task can be stated within the language of our framework. In Sections \ref{sec:learning_to_transfer} and \ref{sec:structure_of_learning_to_transfer}, we introduce our framework, and detail its components respectively. We will then exposit an in-depth discussion that will elaborate the breadth and impact of our framework in the final sections.  
% To this end, having discussed an intuitive introduction to the key ideas of learning to transfer in Section \ref{sec:intuitive_definition}, we will provide a formal mathematical treatment in Section \ref{sec:formal_definition}. The remaining sections will apply this framework to known sub-fields of ML and a few examples of existing methods that are ubiquitous in the literature. 

\section{Learning a Single Task} \label{sec:learning_a_single_task}
At the lowest level of a learning to transfer problem, there is the problem of learning a single task. We begin our explanation of the framework by describing this familiar learning paradigm.
\subsection{Learning Task} \label{sec:learning_task}
A learning task is meant to encompass what we generally think of as a single learning problem in machine learning. This can include supervised learning, unsupervised learning, reinforcement learning (RL), among others. While traditional definitions describe a learning task in terms of generative distributions, we choose an equivalent definition that focusses on structure:
\begin{definition}[Learning Task] \label{def:learning_task}
	A learning task is a single machine learning problem. The problem will consist of some structure $\structure$, which can be used by a generative process $\process: \structure \rightarrow \mathcal{F}_\dataset$ to generate a dataset $D \in \dataset$ via $f \in \process(\structure)$. $(f \in \mathcal{F}_\dataset): \Omega \times \mathcal{Z}^+ \rightarrow \dataset$ is a map that takes a positive integer $n$ and some other inputs $\Omega$ to return the dataset of size $n$. The learning task is a subset of $\structure$ that we would like to learn. We will denote a learning task by $t$.
\end{definition}
In this definition, the generative process is the \emph{mechanism} by which data can be generated; it defines the class of problems we are dealing with. For example, in supervised learning, this process is the application of a generative function; in RL the process involves rolling out the full Markov Decision Process (MDP) that samples a trajectory given the system's transition dynamics, reward function and policy.

The structure $\structure$ then contains everything else that is required to specify the particular instance of the relevent generative process from a set of all possible such processes. As we will describe in Section \ref{sec:ml_structure}, $\structure$ is often a set of maps that satisfy particular properties; the learning task then, is the subset of such maps and their properties that we want to learn,  while the rest can be thought of as an inductive bias that is assumed. The structure of a problem can be described in different ways; a useful description of structure is in terms of a hierarchy of biases that, at each level, creates smaller sets from the set of \emph{all} things we want to consider, until we are able to uniquely identify the task. 

To see this, consider carrying out supervised learning on data generated from a sinusoid $y(x) = \sin(x)$. In writing the previous statement, we have already described the structure of the problem; it is a function $y: \R \rightarrow \R$ that can be written as $y=\sin(x)$. Alternatively, we could also describe $y=\sin(x)$ as a continuous, odd, scalar valued, circular function with domain $\R$ that is periodic with period $2\pi$, and has a constant amplitude of $1$. We can rewrite this as a hierarchy of subsets, starting from the largest, set of maps $\rightarrow$ maps on $\R$ $\rightarrow$ scalar valued maps $\rightarrow$ continous functions $\rightarrow$ odd functions $\rightarrow$ circular functions $\rightarrow$ constant period of $2\pi$ $\rightarrow$ constant amplitude of $1$. Note that each subsequent level inherits the properties of the previous level. We will see in Section \ref{sec:relatedness} that writing structure in this way is crucial to transfer.
\subsection{A Space of Models} \label{sec:models}
Machine learning often involves finding a representation of an aspect of the generative process described above (see Section \ref{sec:representations}). Thus, in order to carry out learning, we need to define a model space from which to choose a suitable representation. The model space can be thought of as a way of representing the task space, and the structure that it carries. In the present work, we will consider parametric model spaces:
\begin{definition}[Parametric Model Space]
	Given an input space $\X$, an output space $\Y$ and a parameter space $\Theta$, a model architecture is given by $\phi: \X \times \Theta \rightarrow \Y$. A parametric model space $\pmodel = \{\phi(\theta, \cdot)\}$,  $\forall \theta \in \Theta$. 
\end{definition}
The parameteric model space $\pmodel$ is therefore a collection of functions that can be indexed by the parameter space, under a given model architecture. Note that the model architecture is a structure that applies to the entire model space; in choosing an appropriate architecture, we have implicitly assumed that the learning task $t$ at hand also carries such a structure. 

The parametric model space $\pmodel$ is isomorphic to its parameter space $\Theta$, which is assumed to be $\Rd$. We will assume that $\Theta$ is a smooth manifold. It should be noted that there is an identifiability issue here; there can be several elements of $\pmodel$ that are functionally equivalent, but with distinct parameters. For example, suppose that $\phi(\theta, x) = \sin(x + \theta)$; due to the periodicity of the sinusoid, we know that $\phi(\theta, x) = \phi(\theta + 2\pi n)$, $\forall x \in \R$ and $n \in \mathbb{Z}$, the set of integers. We will therefore define a model space as follows:
\begin{definition}[Model Space]
    Given a parametric model space $\pmodel$, the model space $\model$ is the quotient space defined as $\model = \Theta/\sim$, where $\sim$ defines an equivalence relation given by:
    \begin{equation}
        \theta_1 \sim \theta_2 \iff \phi(x, \theta_1) = \phi(y, \theta_2), \: \forall x \in \X \: \mathrm{and} \: \theta_1, \theta_2 \in \Theta
    \end{equation}
    
    We will denote the model space as $\model$.
\end{definition}
In general, we can consider other equivalences too. For example, two parameters could be equivelent if the loss value they produce is the same; this means that our learning algorithm cannot differentiate between these. Thus, we can define a general model space to be the quotient space w.r.t an equivelence between the learning algorithm. The construction of such a space is beyond the scope of the present work.  In most practical cases, and for the purposes of the present paper, it suffices to look at local regions of the space; that is open neighbourhoods taken from the topology on the manifold, and therefore, local structures, provided that $\model$ can be described as a smooth manifold (which we assume\footnote{proving that the quotient space described here is a manifold is beyond the scope of the present work, and will be left for future work.}).

\subsection{Learning Algorithm} \label{sec:learning_algorithm}
Given a task, and a model space, we can define a learning algorithm as:
\begin{definition}[Learning Algorithm] \label{def:learning_algorithm}
	Given a model space $\model$, a task space $\task$, and a set of \emph{relevant algorithmic properties} $\Omega$, the learning algorithm $\learning$ is a map $\learning: \{t\} \times \Omega \rightarrow \model$, such that $\learning(t, \omega) = m_t^*$ for some set of algorithm parameters $\omega \in \Omega$, is the \emph{most suitable} model in $\model$ for the learning task $t \in \task$.
\end{definition}
The learning algorithm $\learning$ is the mechanism by which we find a suitable representation of the learning task. Often in machine learning, we measure the suitability of a model for a particular task using a loss function, defined as:
\begin{definition}[Loss Function] \label{def:loss_function_single}
	Given a learning task $t$, and model space $\model$, a loss function $\loss_t$ is a map $\loss_t: \model \rightarrow [0, \infty]$.
\end{definition}
Note here, that the loss function is defined for a particular learning task $t$. It is often assumed that $\model$ contains at least a single model $m_t^*$ such that $\loss_t(m_t^*) < \epsilon \in \R^+$, for a desired accuracy $\epsilon$. The learning algorithm is a map from a learning task to the most suitable model, in this sense. The set of algorithmic parameters are hyperparameters that define the learning algorithm. The loss function can be included in this set, as well as the optimiser that is used, to name another example.

\subsection{The Problem of Learning a Single Task}
\begin{figure}[ht]
    \centering
    \includegraphics[width=0.6\textwidth]{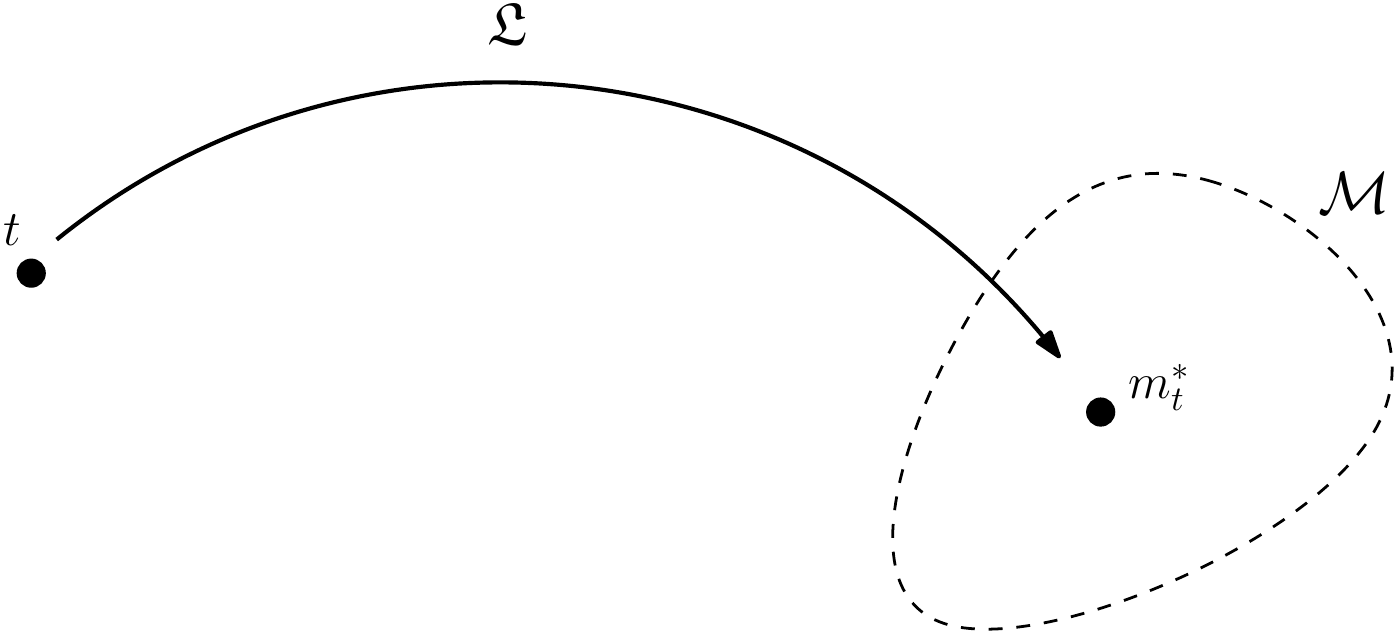}
    \caption{Learning a single task. Here, $m^*_t$ represents the optimal solution to the learning task.}
    \label{fig:learning_a_single_task}
\end{figure}

Putting these altogether, we make the following definition:
\begin{definition}[The Problem of Solving a Single Learning Task] \label{def:single_task_learning}
    Suppose we are given a learning task $t$. The problem of solving a single learning task is to 
    \begin{enumerate}
        \item construct the model space $\model$ that contains a sufficiently suitable description of the structure defining $t$. 
        \item construct a loss function $\loss_t: \model \rightarrow [0, \infty]$ that appropriately measures the suitability of $m \in \model$ as a description of $t$,
        \item and devise a learning algorithm $\learning$ that maps $t$ to $m_t^* = \argmin_{m \in \model}\loss_t(m)$ is the most suitable model in $\model$, according to $\loss$.
    \end{enumerate}
\end{definition}
A graphical representation of this problem is shown in Figure \ref{fig:learning_a_single_task}.
%%%%%%%%%%%%%%%%%%%%%%%%%%%%%%%%%%%%%%%%%%%%%%%%%%%%%%%%%%%%%%%%%%%%%%%%%%%%%%%%%%%%%%%%%%%%%%%%%%%%%%%%%%%%%%%%%%
\section{Learning to Transfer} \label{sec:learning_to_transfer}
% The foliated theory of learning to transfer describes its namesake problem using the following steps:
% \begin{enumerate}
%     \item definition of a \emph{learning task}
%     \item definition of a \emph{task space}, a \emph{model space}
%     \item definition of, and distinction between \emph{relatedness} and \emph{similarity}
%     \item definition of an \emph{equivariant learning algorithm} w.r.t to the chosen relationship between tasks.
% \end{enumerate}

\begin{figure}[ht]
    \centering
    \includegraphics[width=0.7\textwidth]{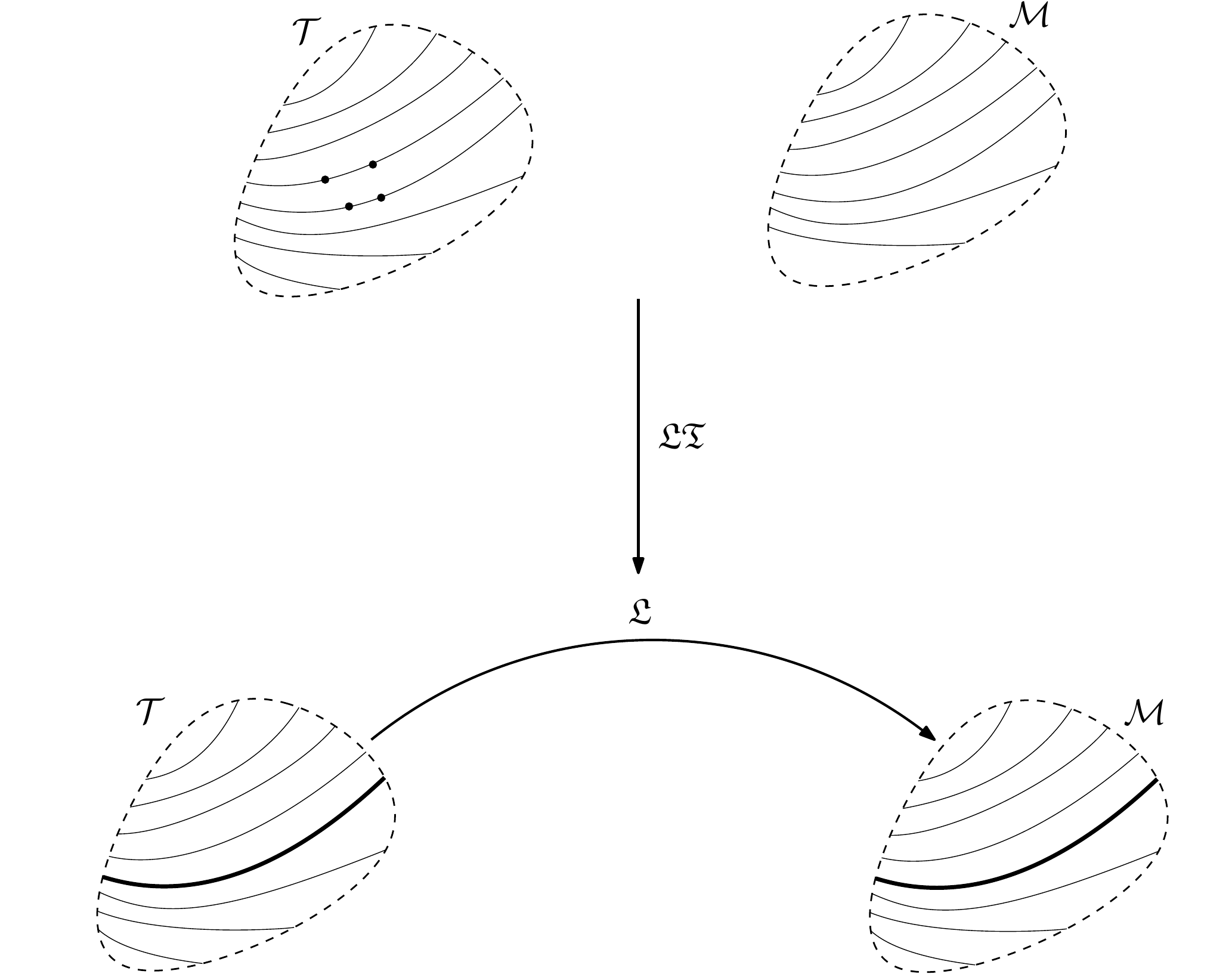}
    \caption{Learning to transfer. Here, we have chosen a notion of relatedness, as made precise in Section \ref{sec:relatedness}. Here, we assume that we are initially given some tasks, denoted by the black dots; we choose a notion of relatedness that applies to both $\task$ and $\model$. The learning to transfer algorithm $\learningtransfer$ then generates a learning algorithm $\learning$ that is equivariant w.r.t. the chosen notion of relatedness.}
    \label{fig:learning_to_transfer}
\end{figure}

% The framework will have the following properties or features, in no particular order, that make it desirable: 
% \begin{itemize}
%     \item it will start from very basic mathematical objects, and add additional structure as needed. In this way, our framework is fundamental. 
%     \item it will treat the learning to transfer problem as a meta-description of the standard learning problem. That is, we posit that transfer is possible at different levels of a hierachy, and the learning to transfer which we consider looks at the first level (assuming zero indexing).
%     \item learning to transfer will be described as a finding structure in the form of learning hierarchical bias. This requires us to look at ML as learning structure; we will discuss this in Section \todo[inline]{add link to section here.}
% \end{itemize}
The formal definition of learning to transfer, as defined within our framework is as follows; we will elaborate on its many components subsequently: 
\begin{definition}[The Problem of Learning to Transfer] \label{def:transfer_learning}
Suppose there is a space of learning tasks $\task$, represented a smooth manifold. From a subset of tasks $U_\task$, we are given a finite number of learning tasks. Using these, the problem of learning to transfer is to
\begin{enumerate}
    \item construct the model space $\model$ that contains sufficiently suitable descriptions of $U_\task$, 
    \item choose a notion of (or a class of) $\Pi_\task$-relatedness on $\task$ and the corresponding, homomorphic $\Pi_\model$ on $\model$,
    \item and devise a learning to transfer algorithm $\learningtransfer$ that produces a learning algorithm $\learning$ that is equivariant w.r.t. the chosen, or \emph{in some sense} optimal $\Pi_\task$,
\end{enumerate}
such that they allow for biased learning of subsequent tasks $t_{\scriptstyle{new}} \in U_\task$. 
\end{definition}
This definition is quite general. It has several components that allow for choice by the designer of a learning to transfer system. In particular, it allows us to encompass the many different ways in which the literature has approached transfer. See Section \ref{sec:relation_to_existing_work} for more details. A graphical depiction of this is shown in Figure \ref{fig:learning_to_transfer}.
%%%%%%%%%%%%%%%%%%%%%%%%%%%%%%%%%%%%%%%%%%%%%%%%%%%%%%%%%%%%%%%%%%%%%%%%%%%%%%%%%%%%%%%%%%%%%%%%%%%%%%%%%%%%%%%%%%
\section{The Structure of Learning to Transfer} \label{sec:structure_of_learning_to_transfer}
We can now describe our framework for learning to transfer.

\subsection{The Space of Learning Tasks} \label{sec:space_of_learning_tasks}
If we compare Definition \ref{def:single_task_learning} of the problem of learning a single task, to that of learning to transfer, as given by Definition \ref{def:transfer_learning}, an important difference is that in the latter, we look at spaces of tasks, rather than just a single task.
\begin{definition}[Task Space]
	A task space is a collection of unique learning tasks. 
\end{definition}
A task space simply contains learning tasks. In particular, this space is simply a set that does not have any other useful structure that is attached to it, other than the tasks that it contains. In this subsection, we argue that it is useful to think about the task space as a smooth manifold. Formal definitions of what this is will be given in the Appendices.

A topological manifold  \citep{Lee2001} is a generalisation of a topological Euclidean space; the global topology of a manifold can be different from a Euclidean space, but locally it would look like a subset of one. An example of this is the circle manifold $\Circ^1$, which is a 1-dimensional topological manifold. 

\begin{figure}
    \centering
    \includegraphics[width=0.6\textwidth]{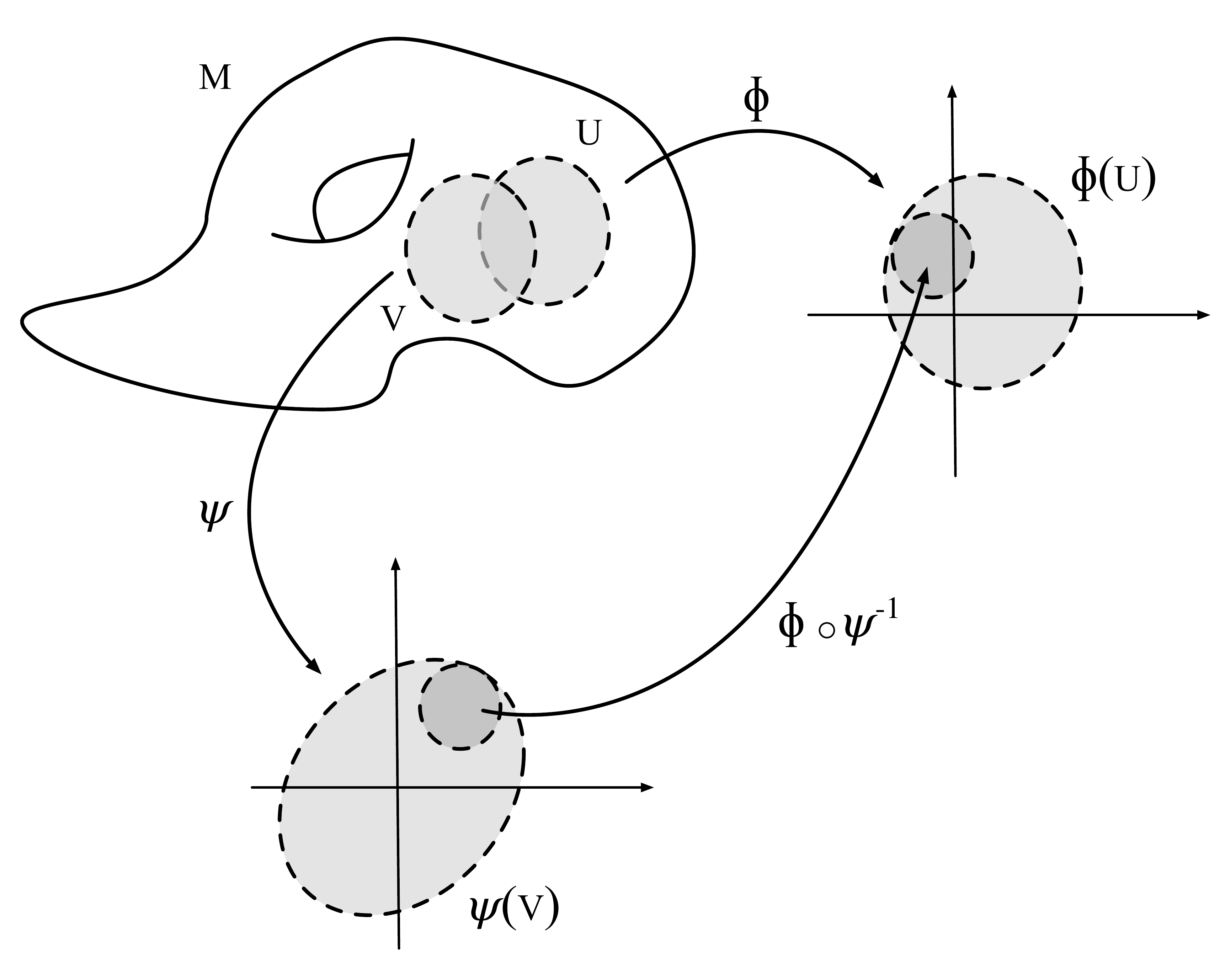}
    
	\caption{A topological manifold $M$ with examples of its coordinate charts. We see here that despite being a complicated topological entity (see the hold in the middle), locally, it is equivalent to $\R^2$. Further, note that the chart transition map $\phi \circ \psi^{-1}: V_{\R^2} \rightarrow U_{\R^2}$ exists because a homeomorphism is invertible and continuous, since the composition of continuous maps remains so.}
	\label{fig:topological_manifold_in}
\end{figure}

An important aspect of a topological manifold are its local coordinates. These are homeomorphisms from local, open neighbourhoods of the manifold on to a fixed dimensional Euclidean space; these maps are called charts, and the collection of charts covering the manifold is called an atlas. Any particular neighbourhood can be given several different, yet valid charts. In fact, there can be regions that intersect with different charts; on a manifold, the map between the charts over the intersected region is also homeomorphism (see Figure \ref{fig:topological_manifold_in}).

These charts can be thought of as a representation scheme (see Section \ref{sec:representations} for more details) of the tasks. The dimensionality of the manifold describes the number of degrees of freedom we have to move in to describe tasks in the constrained set of tasks $\task$. This is opposed to a universal set of tasks that contains \emph{all possible} learning tasks; we believe that this is too general to consider\footnote{There does exist an obvious counter example to this however: a reasonable set of tasks that one could consider in a regression task is the set of all continuous functions. This is technically \emph{infinite} dimensional. While we believe that it is possible that our theory could be extended to this case, we will restrict ourselves to the finite dimensional scenario for the time being.}. The charts then describe local coordinate systems that \emph{quantify} these degrees of freedom. However, the particular chosen charts themselves do not matter. Theories on manifolds are coordinate independent. 

This is useful for thinking about task spaces and learning tasks, since we do not want to have to worry about how to represent a task. We want to consider tasks as abstract objects, that exist with particular structure. This also extends to the full space of tasks. The manifold structure allows us to think about this.

The finiteness of the task space could be thought of as pathological; for example, it is reasonable to consider the space of all continuous functions as a task space. However, we propose that given some useful structure, and therefore bias, we will often find subspaces of this full set which are indeed finite dimensional, and form a topological manifold. Certainly, it is possible to find such subspaces, and to therefore restrict our attention to solving learning tasks from such spaces. We mentioned in Section \ref{sec:learning_task} that the elements of a task space themselves provide the most fundamental level of structure, since they are chosen, from the set of all possible tasks to belong to the given task space $\task$. For example, if we define the task space to be all regression tasks generated from scalar valued truncated Fourier series on $\R$, we immediately note there is a constraint (and therefore structure) that defines the task space.

Finally, a topological manifold, as the name would suggest, must contain a topology on the set of things on the manifold. A topology gives us a concrete notion of neighbourhoods of elements; elements that are close to each other\footnote{It should be noted that while we used the term \emph{close} here, a topology does not require a metric, or a notion of distance, although a metric topology can be defined from one. A topology merely tells us about which points are near each other.}. A topology is a very basic structure that is placed on a set of things, in this case tasks. It allows us to start organising our space of tasks.

To be a topological manifold, $\task$ requires there be a topology on the task space. In our framework, we would assume a reasonable topology on the task space, where neighbourhoods are formed by tasks we expect to be close to similar to each other. However, in practice, specifying a loss function and model space can generate a topology on the task space. For example, we could choose any topology for which the loss function is continuous w.r.t. the task space. An example of such a topology is the following:
\begin{theorem}[Induced Topology on Task Space] \label{theo:topology}
	We are given a task space $\task$, a model space $\model$, a loss function $L: \task \times \model \rightarrow [0, \infty]$ and a learning algorithm $\learning_L: \task \rightarrow \model$. Assume that each $L_{t \in \task}: \model \rightarrow [0, \infty]$ is continuous. Then, a loss ball centered at $t \in \task$ is given by 
	\begin{equation}
		\mathcal{B}_t(\epsilon) = \{s \in \task : \|L(s, \learning(t) ) - L(t, \learning(t))\| < \epsilon\}
	\end{equation}
	for $\epsilon > 0$.
	A subset $U_\task \subseteq \task$ is then an element of the induced topology $\mathcal{O}_\task$ iff \ $\forall t \in U_\task, \: \exists \: \epsilon > 0$ s.t. $\mathcal{B}_t(\epsilon) \subseteq U_\task$.
\end{theorem}
The proof that this is a valid topology can be found in Appendix \ref{sec:topology_proof}. Note that the definition of a loss function has been extended compared to Definition \ref{def:loss_function_single}, and as far as the rest of the present work is concerned, redefined, as the following:
\begin{definition}[Loss Function]
	Given a learning task $t$ which comes from a task space $\task$ and model space $\model$, a loss function $\loss$ is a map $\loss: \model \times \task \rightarrow [0, \infty]$.	
\end{definition}
Theorem \ref{theo:topology} defines a topology similar to a metric topology, but defined w.r.t. a loss function. Here, the open sets of the topology consist of tasks that are close to each other given the loss function. 

A topological manifold endowed with a smooth structure is called a smooth manifold. This is a requirement that all chart transition maps be smooth. Such a structure allows us to talk about notions of calculus on a manifold. These include gradients, vector fields, flows, integration, and many others. As such, we can think of a task space as a smooth manifold, since notions of differentiation are useful in ML. For example, to think about gradient descent is to think about flows along a manifold. 
\subsection{Relatedness} \label{sec:relatedness}
In Definition \ref{def:learning_task}, we made specific mention of structure. The elements that constitute the task space themselves inform us of a fundamental structure that identifies this particular set of tasks from the set of all possible tasks. Knowledge of such constraints would immediately allow us to meaningfully transfer between elements of this task space. In practice however, we lack access to such knowledge, and learning to transfer amounts to learning the structure that defines a subset of a large space of tasks. That is, given a rather large space of tasks, the structure of which we know of, we want to find a hierarchy of this space that creates subsets of it, each of which carries some inductive bias, or structure that defines it, within the original space of tasks. 

Each level of structure describes some inductive bias which identifies a smaller subset of tasks. We posit that transfer can be then written as identifying the bias that describes a subset of tasks, and exploiting this bias to making learning within this subset more efficient. The efficiency gains can immediately be attributed to the smaller size of each subset. There is not a unique hierarchy of biases that can be used to describe a learning task. The choice of such a hierarchy depends on the needs of the learned system (we will discuss this more in Section \ref{sec:choosing_foliation}).

\begin{figure}[ht]
    \centering
    \includegraphics[width=0.7\textwidth]{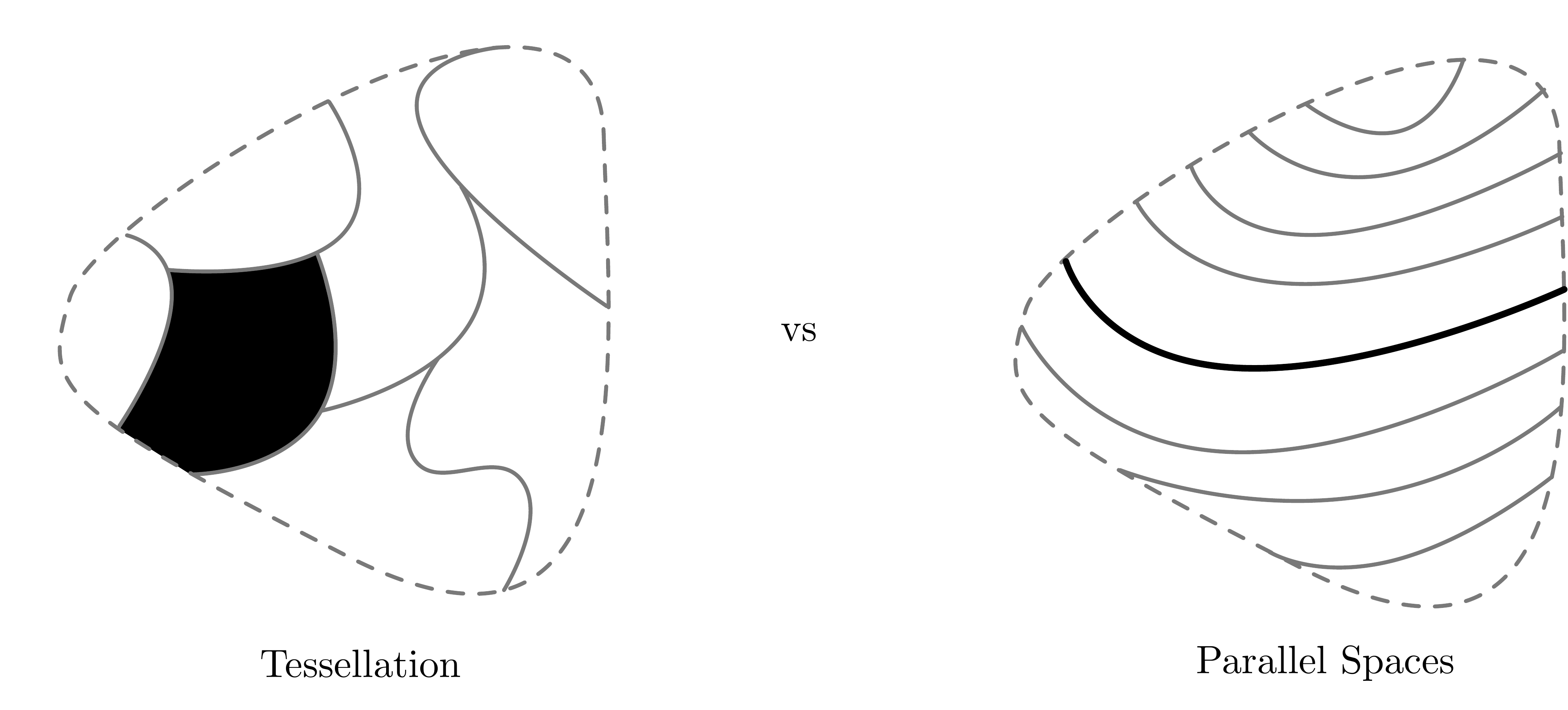}
    \caption{Comparison of tessellation and parallel spaces. The shaded regions denote the subsets we are considering in each case. A tessellation creates disjoint, smaller subsets that are of the same dimension as the original set. Parallel spaces on the other hand creates disjoint subsets that are of a lower dimension than the original set.}
    \label{fig:tessellation}
\end{figure}

Learning to transfer at a chosen level will then involve exploiting this hierarchy at that level. For this, we are required to define a scheme that can be used to describe such a hierarchy. We propose two methods that will lead to a precise notion of relatedness. Consider a space of learning tasks $\task$. A hierarchy with a single level can be described as a partition of $\task$. The hierarchy then consists of the identities of each partition at the highest level, then the learning tasks that belong to each partition form the lowest level. Then, 
\begin{enumerate}[label=\alph*)]
    \item we can tessellate $\task$, where we create subsets (or submanifolds) that are of the same dimension as $\task$. 
    \item we can create parallel submanifolds that are of dimensions lower than $\task$. 
\end{enumerate}
Note that the dimensions of the submanifolds need not be equal to each other. Figure \ref{fig:tessellation} graphically shows examples of these. It is possible to consider hierarchies that consist of combinations of these partitions.

A key aspect of either partition is that each subset defines an equivalence relation that uniquely identifies the membership of a particular task in its partition. That is, given a partition of $\task$, $\{U_i\}_{i \in \index}$, then for $t_1, t_2 \in \task$, $t_1 \sim t_2 \iff t_1, t_2 \in U_i$. Thus, we can define a notion of sets of \emph{related} tasks as,
\begin{definition}[Sets of Related Tasks] \label{def:sets_of_related_tasks}
    Given a set of tasks $\task$, and partition $\partition_\task = \{U_i\}_{i \in \index}$, each $U_i \subseteq \task$ is defined as a set of related tasks.
\end{definition}
Then, 
\begin{definition}[A Notion of Transfer]
    Given a set of tasks $\task$, a partition $\partition_\task = \{U_i\}_{i \in \index}$ on $\task$ will be called a \emph{notion of transfer}.
\end{definition}
Such a notion of transfer allows us to define transfer as a transformation. Each subset $U_i \in \partition_\task$ is naturally equipped with a permutation group of transformations that allows us to transform one task to another within a set of transferable tasks. In the case when $\task$ is discrete and finite, the permutation group is also discrete and finite; in other cases, such as when $\task$ can be written as a smooth manifold, this group can be quite unwieldy. In general, it might be sufficient to consider smaller set of pseudogroups that reproduce $\partition_\task$. Such a set of pseudogroups will be called a notion of \emph{relatedness}.

\begin{definition}[A Notion of Relatedness]
    Given a set of learning tasks $\task$, a set of groups or pseudogroups $\Pi_\task = \{\group_i\}_{i \in \index}$ of transformations which act on subsets of $\task$ is called a \emph{notion of relatedness} if:
    \begin{enumerate}
        \item $U_i, U_j$ are the domains of $\group_i, \group_j \in \Pi_\task$, then $U_i \cap U_j = \emptyset$,
        \item $\bigcup_{i \in \index}{U_i} = \task$,
        \item $\bigcup_{i \in \index}\orbit_i$, where $\orbit_i$ is the set of orbit $\group_i$, is a partition of $\task$. An element of such a partition is then \emph{a set of related tasks}.
    \end{enumerate}
\end{definition}
Definitions of groups and pseudogroups are given in Appendix \ref{app:groups} and Appendix \ref{app:pseudogroups}, respectively. Then, we can define transfer as being:
\begin{definition}[Transfer]
    Given a notion of relatedness $\Pi_\task$, \emph{transfer} from $t_1 \in U_i \subseteq \task$ to $t_2 \in U_i \subseteq \task$ is the action of a transformation from $\group_i \in \Pi_\task$.
\end{definition}

Thus, transfer can be precisely thought of as transformations between elements in a set of related tasks. That is, two tasks are related if they exists in a set of related tasks. One can induce the sets of related tasks by first defining a notion of relatedness, or conversely, induce relatedness via a defined partition of $\task$. $\Pi_\task$ was chosen to contain sets of groups and/or pseudogroups for reasons of consistency; in particular, such restrictions allow transformations to be agnostic of the order of transformations taking place. They also ensure that inverse and identity transformations exist, making our set of transformations intellectually pleasing and interpretable.

Our notion of relatedness is similar to a definition given in \citep{Ben2003, Ben2008}, where a single group of transformations act on $\task$; our definition generalises this notion to include different groups of transformations that can act on subset of $\task$, but produce the partition as needed. 

It also agrees with what we intuitively expect relatedness to mean. In our introductory example, we wanted to transfer from one pendulum to another; the two pendulums are \emph{related} by the fact that they are both pendulums. We could select a mass and length from $\R^+$, and still obtain what we would classically think of as a pendulum, by plugging them into the differential equations of a pendulum. With relatedness, we want to capture the notion that there is a set of properties that distinguish a set of things, and that there is set of transformations that preserves these properties; one can think of these as factors of variation that preserve the identify of a set of related things. Such a set of properties are the inductive biases we had mentioned previously. Here, these properties are the information encoded in the differential equations of a pendulum. In Section \ref{sec:invariances}, using a particular notion of relatedness, we can show that there are invariant quantities that can be thought of as identifying structure of a set of related things.

In particular, knowledge of the invariant properties of a set of related tasks allows us to identify the tasks in such a set using less information than if we were to identify them from $\task$; we can assume the knowledge that is given by the membership of a task in a particular set of related tasks. In this way, the key benefit of transfer can be seen. A notion of relatedness provides us with a smaller set of factors of variation to consider when describing tasks. Another way to see this is to consider that it is possible to write a single set of transformations that acts transitively on $\task$. Here, all tasks are related to each other. If we trivially consider this set to be the group of permutations of $\task$, then, we can note that such a group would be larger than the group of permutations of a subset of $\task$. 
\subsection{Representing Relatedness Using Foliations}
A foliation is a geometric structure that can be placed on a smooth manifold\footnote{Smoothness is not a necessary condition to introduce a foliation on a manifold, but we consider it for our present purposes.}. In particular, a foliation can allow us to mathematically write partitions such as those we saw in Figure \ref{fig:tessellation} on a smooth manifold. 
\subsubsection{Foliations} 
A foliation is a restriction on the atlas that is given to a smooth manifold. There are two types of foliations.
\begin{definition}[Regular Foliation] \label{def:foliation}
	If $M$ is a smooth manifold of dimension $d$ with an atlas $\A_\model$, a $n$-dimensional foliation $\mathcal{F}$ of $M$ is a subset of its atlas, that satisfies:
	\begin{enumerate}[label=\alph*)] 
		\item if $(U, \phi) \in \F$, then $\phi: U \rightarrow (U_1 \subseteq \R^{d-n}) \times (U_2 \subseteq \R^n)$, and 
		\item if $(U, \phi), (V, \psi) \in \F$, where $U \cap V \neq \emptyset$, the chart transition functions, such as $\psi \circ \phi^{-1}: \phi(U \cap V) \rightarrow \psi(U \cap V)$ has the form of $\psi \circ \phi^{-1}(x, y) = (h_1(x), h_2(x, y))$, where $x \in U_1$ and $y \in U_2$.
	\end{enumerate}
\end{definition}
\begin{figure}[ht]
    \centering
    \includegraphics[width=0.6\textwidth]{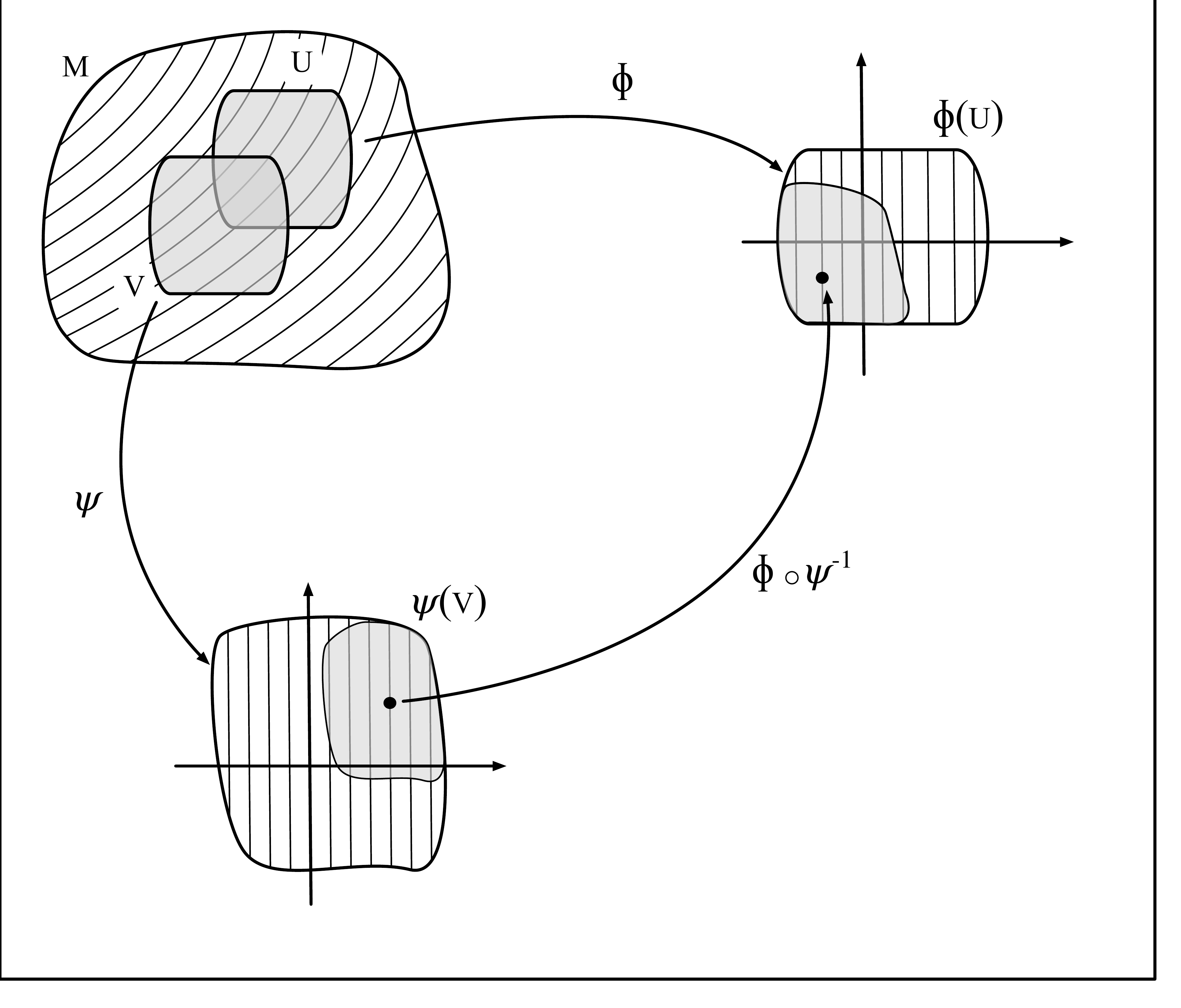}
    
	\caption{A 1-dimensional regular foliation on $\R^2$.}
	\label{fig:foliation}
\end{figure}
Intuitively speaking, this definition describes the decomposition of a manifold into non-overlapping connected submanifolds that are of dimension $n$ \citep{Camacho2013}. Each such submanifold is called a \emph{leaf}.

A key feature of a regular foliation is that the dimension of the leaves is constant. In this case, the leaves create a partition of the manifold that looks similar to how we described parallel spaces. However, a generalisation of this exists, called a singular foliation \citep{Sussman1973, Stefan1974}, which allows us to write tesselations, and combinations of the two styles of partitioning in the same language. 
\begin{definition}[Singular Foliation \citep{Stefan1974}] \label{def:singular_foliation}
	If $M$ is a smooth manifold of dimension $d$ with an atlas $\A_\model$, a singular foliation is a partition of $M$ into immersed, connected submanifolds of non-constant dimensions. In particular, there exists an atlas of \emph{distinguished} charts, where for each $x \in M$ there exists a chart $\phi$ such that:
	\begin{enumerate}[label=\alph*)] 
		\item if $x \in U$, then $\phi: U \rightarrow (U_1 \subseteq \R^{d-n}) \times (U_2 \subseteq \R^n)$, where $n$ is the dimension of the leaf containing $x$, and $U_1$ and $U_2$ contain $0$,
		\item $\phi(x) = (0, 0)$, and  
		\item if $L$ is a leaf, then $L \cap \phi^{-1}(U_1 \times U_2) = \phi^{-1}(l \times U_2)$, where $l = \{w \in U_1: \phi^{-1}(w, 0) \in L\}$.
	\end{enumerate}
\end{definition}
It should be noted that foliations, in general, can have complicated topologies and properties on their leaves, and space of leaves \citep{Rosenberg1970, Camacho2013, Lawson1974}.
\subsubsection{Relevance to transfer}
Suppose we are given a task space $\task$, which we write as a finite dimensional smooth manifold. We can then define a notion of transfer on this by defining a foliated structure $\foliation$, regular or otherwise, on $\task$. The following theorem shows that, given any foliation, we can construct a set of pseudogroups of transformations on $\task$ that can behave as a notion of relatedness.
\begin{theorem}[Relatedness from $\foliation$] \label{theo:relatedness_foliation}
    Given a leaf  $L$ of a foliation $\foliation$ on a manifold $\task$, there exists a (pseudo)group of transformations that act transitively on $L$. Then the collection of such (pseudo)groups over all leaves of $\foliation$ is a notion of relatedness.
\end{theorem}
A proof of this is given in Appendix \ref{sec:relatedness_foliation}.

The converse is also possible, where we define a notion of relatedness, and derive a foliation from it. Foliations, particular singular foliations can be generated in many ways. For example, this is of great importance to the field of geometric control theory, where singular foliations are used to describe solvable families of control strategies \citep{Jurdjevic1997}. Several results in the study of singular foliations \citep{Lavau2018} were derived for this; in particular, it has been shown in an Orbit Theorem \citep{Jurdjevic1997}, that orbits of families of vector fields, under certain conditions generate singular foliations. Since we can use such families of vector fields to generate (local) diffeomorphisms, they can give us a notion of relatedness. More simply however, it is known that the locally free action of a Lie group produces a regular foliation \citep{Camacho2013,Lawson1971}. 

Thus, using foliations, we can go either way. If a notion of relatedness that is thought to be relevant to a particular problem of learning to transfer is known, we can apply it to generate a structure on a task space that reflects the induced notion of transfer. Alternatively, perhaps, a foliation can be learned, or simply defined, from which a notion of relatedness can be derived, for purposes of interpretability and understanding. 
\subsection{Equivariant Learning Algorithms} 
So far, we have defined a notion of relatedness on $\task$. Since we want the model space to reflect our assumptions about $\task$ (see Section \ref{sec:representations}), we would want there to be a notion of relatedness that is \emph{similar}. Here, we define this similarity to be group, or pseudogroup homomorphisms, given the notion of relatedness. That is, given $\Pi_\task$, we would want $\Pi_\model$ to contain elements that are homomorphic to the elements of $\Pi_\task$; the simplest way to achieve this is to assume that they are the same, and that $\task$ and $\model$ are also the same. 

Following this, it is natural to expect that an appropriate learning algorithm, which is able to transfer, ensures that sets of related tasks are mapped to sets of related models (for this, simply replace $\task$ with $\model$ in Definition \ref{def:sets_of_related_tasks}). This brings us to the final part of Definition \ref{def:transfer_learning} --- a learning to transfer algorithm $\learningtransfer$ produces an equivariant learning algorithm $\learning$. $\learning$ here is as defined in Section \ref{sec:learning_algorithm}, but extended to work on $\task$, rather than a single task.
\begin{definition}[Equivariance]
    Given a group $\G_M$ and an action $\theta_M$ on a space $M$, a group $\G_N$ and an action $\theta_N$ on a space $N$, and a group homomorphism $\rho: \G_M \rightarrow \G_N$, we say that a map $f: M \rightarrow N$ is equivariant if $f(\theta_M(g_m, m)) = \rho(g_m)(f(m))$, for any $m \in M$ and $g_m \in \G_M$. 
\end{definition}
Equivariance \citep{Olver1995} is a property which, not unlike invariance, can be placed on maps, and transformations that act on the domain and codomain. Here, instead of the output staying constant when the domain is transformed, we expect the output to also change in a similar manner. Similar here implies  either that the transformation set is the same, or that there is a structure preserving map to a transformation set that acts on the codomain. This notion has been growing in interest in the ML community; for example, it is known that CNNs are translation equivariant, as visual problems follow similar properties \citep{Cohen2018, Cohen2019}. 

A learning algorithm $\learning$ is equivariant, if given a transformation $\pi \in \Pi_\task$ which acts on $t \in \task$,it it satisfies $\learning(\pi(t1)) = \rho(\pi)\learning(t_1)$, where $\rho(\pi) \in \Pi_\model$. Thus, the goal of a learning to transfer algorithm is to produce a learning algorithm that satisfies this property.

In the most general case, choosing a notion of relatedness \emph{a priori} is infeasible, and we would want to learn the most suitable notion. In this case, we can define a class of notions of relatedness (for example, by parameterising foliations), and a learning to transfer algorithm that optimises equivariant learning algorithms. Section \ref{sec:choosing_foliation} discusses some preliminary ways by which to measure suitability of a notion of relatedness.

% !TEX root =  /home/janith/dev/jmlr_transfer_learning/v0.tex
\section{Relation to Existing Work} \label{sec:relation_to_existing_work}

In this section, we will show that a lot of work in the literature can come under the umbrella of our framework; in the following section, we will show this fit more precisely for a few archetypal examples. We will see that several problem statements that come with different names describe the same underlying problem. This problem is what we have described previously, the one of trying to find structure within a set of tasks. We will see that the solution that are sought  in fields such as meta learning, transfer learning, continual learning, and the like, can be cast as specific instances under different assumptions of the key variables of the framework. We will begin with Meta-Learning.

In Definition \ref{def:transfer_learning}, we stated that learning to transfer is to equivariantly learn a foliated structure on the model space; this foliated structure will be chosen to be suitable for some set of tasks that are trained on. The foliated structure corresponded to having learnt an inductive bias that informs us about a set of related tasks that our tasks belong to. Knowledge about the set of related tasks to which our tasks belong to then allows us to \emph{assume} this bias in subsequent tasks that are taken from the same set of related tasks; this makes learning faster, or easier. 

There are several scenarios that we can imagine that this general setup can generate. In the first instance, let us look at when the set of tasks that are used at training are available \emph{simultaneously}. This corresponds to the type of problem that is tackled by multitask learning \citep{Caruana1997}, transfer learning \citep{Pan2009,Baxter2000}, meta-learning \citep{Hospedales2020,Finn2017} and few-show learning \citep{Snell2017,Sung2018}. The key differences between these boils down to their use case and learning algorithms. The types of models that each of these techiques use can be very similar. We can also further differentiate these methods in terms of whether they use use a notion of similarity or relatedness to make learning faster. 

\begin{figure}[t]
	\centering
	\begin{subfigure}[t]{0.45\textwidth}
		\centering
		\includegraphics[width=0.6\textwidth]{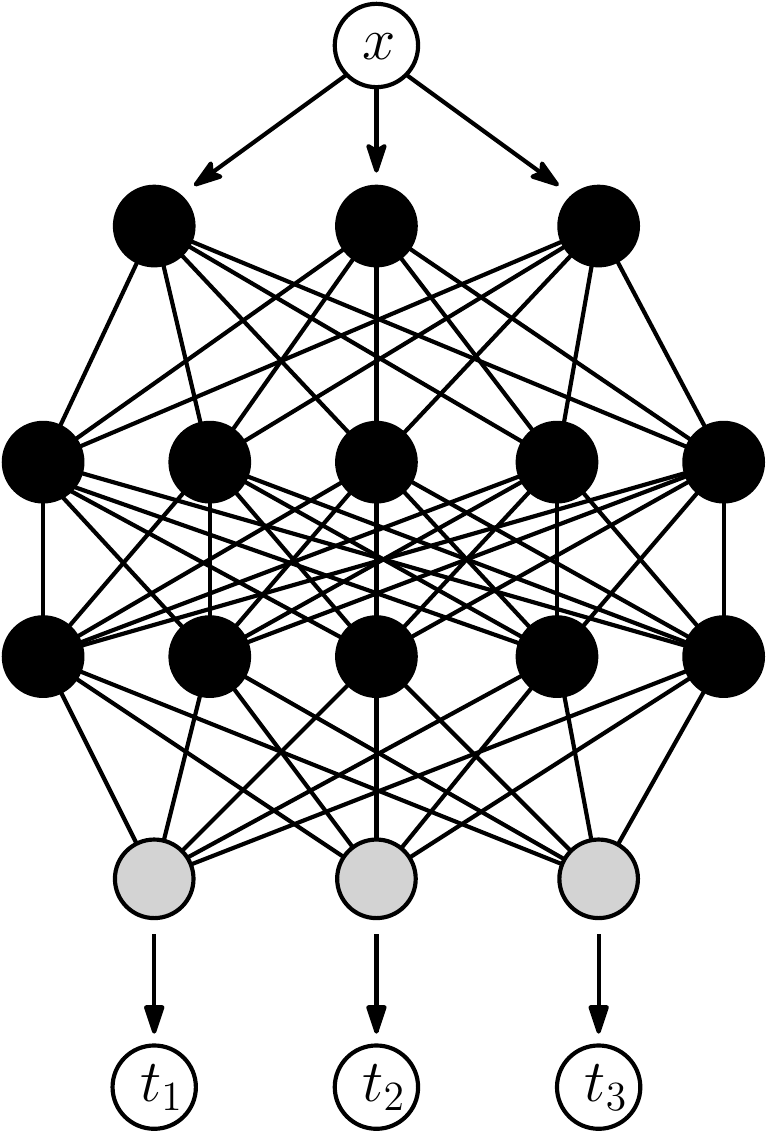}
		\caption{Hard parameter sharing in multitask learning using a neural network. The black nodes represent nodes that are shared between all tasks, whereas the gray nodes are task specific.}
		\label{fig:mt_learning}
	\end{subfigure}
	\hfill
	\begin{subfigure}[t]{0.45\textwidth}
		 \centering
		\includegraphics[width=0.6\textwidth]{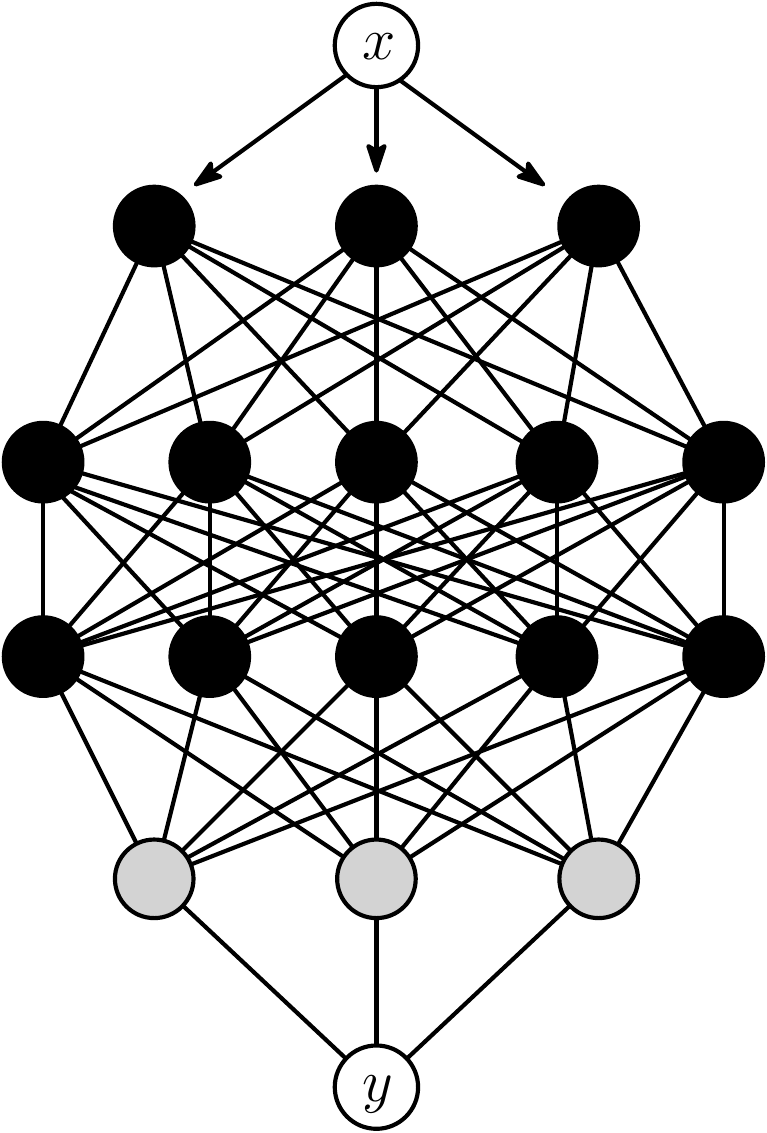}
		\caption{Weight sharing in inductive transfer learning. The black nodes represent nodes that are fixed during learning the new task, whereas the gray nodes are updated during this time.}
		\label{fig:transfer_learning}
	\end{subfigure}
    
	\caption{Example network architectures for multitask learning and inductive transfer learning.}
\end{figure}

Multitask learning uses the relatedness structure between a set of tasks to learn \emph{better} than if we learned on just a single task. We can imagine this as reusing data; if given 2 tasks that are related to each other, we can assume that the data that they generate would also be related. That is, the data from $t_1$ can be \emph{useful} to solving $t_2$, but maybe not as much as data from $t_2$ itself\footnote{Super transfer \citep{Hanneke2019} is an interesting phenomenon, where data from $t_1$ is more useful for $t_2$ than data from $t_2$ itself.}. Thus, we could use data from $t_1$ to augment the learning of $t_2$, and vice versa. This is usually done via some form of parameter sharing, either hard (shown in Figure \ref{fig:mt_learning} or soft; these models in fact correspond to the local representation of a foliation. In multitask learning, we do not assume that we see a subsequent task. The correspondence with our framework should be clear here. 

Transfer learning, particularly that done via inductive transfer \citep{Pan2009, Baxter2000} does. Typically, we initially train a neural network\footnote{We describe transfer learning in neural networks as they are the most clear, pedagogically. However, transfer learning is not restricted to neural networks \citep{Skolidis2012}.} on a vast dataset that contains information about several different problems; these problems could be defined as being separate tasks. We would then be given a new dataset from a particular task, and be expected to adapt the learned model to this new task. This is often carried out by fixing several layers of the network, and retraining the remaining. This is depicted in Figure \ref{fig:transfer_learning}. The weights of the layers that are learnt initially then correspond to the bias; it contains information that is relevant to the current task, and the updated parameters contain the task specific information, and must be learned again. 

According to \cite{Pan2009}, transfer learning can also be transductive. In the previous case, the domain and codamains of the tasks remained constant, whereas the maps between them changed. They called this a change in the task, but our definition of the task encompassed the domain and codomain too. Transductive transfer could involve, for example, a change in the domain; this is called domain adaptation. Typical solutions in domain adptation involve learning intermediate representation maps that can be used on subsequent tasks, simiilar to inductive transfer learning \citep{Wang2018}; the key differences lie in the losses used, and perhaps other intuitive constraints, such as geometric ones \citep{Chopra2013}.

Meta learning, or learning to learn is a recently revitalised field that is popular. Historically, it was initially discussed in papers such as \cite{Thrun1994}, \cite{Schmidhuber1996} and \cite{Thrun2012}. \cite{Schmidhuber1995} and \cite{Schmidhuber1996} introduced a lot of the intuition that we have tried to expound in the present work, but focussed on a machine perspective of the system. Schmidhuber really identified how most of what we think about when trying to learn across several tasks boiled down to learning biases over these tasks. More modern methods in this field, such as MAML \citep{Finn2017}, Siamese networks \citep{Bertinetto2016} and CAVIA \citep{Zintgraf2019} are deep learning based methods that follow the principles of learning that are familiar today.

Meta learning \citep{Hospedales2020} under the deep learning paradigm can be described as a multitask learning paradigm that learns over two optimisation levels; the higher level which learns something that is useful across all tasks, and the lower level that optimises task specific parameters. During the inference phase, where a new task is provided, assumed from the same set of related tasks, only the lower level optimisation is carried out. This should sound familiar at this point. The learning to learn paradigm involves learning what is useful for the given set of tasks, as is done by the higher level optimiser; in practice, this could be, for example, choosing hyperparameters, or a network architecture (Neural Architecture Search). Often, few shot learning is conflated with meta learning. 

The second class of methods are applied to tasks that are obtained sequentially. Continual learning \citep{Lesort2020,Lesort2020_1,De2019}, also called lifelong learning \citep{Thrun1995}, incremental learning, sequential learning or online learning \citep{Hoi2018}, assumes that data from different tasks arrive in a sequential fashion. Here, on an episode, the learner is given data from a particular task, and is asked to adapt to this task, without forgetting what has been learned in previous episodes; additional constraints could include a lack of access to past data. Forgetting what has been learned previously is, quite bombastically, called catastrophic forgetting\footnote{We can try to understand catastrophic in terms of similarity (see Section \ref{sec:similarity}); the further you move away from a particular solution, the less similar you become. At some point, resemblence to the previous task has sufficiently vanished, that we can classify the learner to have forgetten this.}. 

These tasks are of course assumed to be related; if not, then catastrophic forgetting is not an issue. The assumption that past tasks are useful to present tasks implies that these tasks are related, that there is something to transfer. As such, we can pose the continual learning problem attempting to learn the bias of a set of related tasks, such that subsequent tasks can exploit this knowledge without forgetting. This is exactly what we had described previously, except for the fact that the tasks appear sequentially, rather than at the same time. 

Curriculum learning \citep{Bengio2009_cur} is an interesting iteration on this sequential learning problem. Inspired by human learning, a learner in this case is given a specifically chosen sequence of tasks, where learning occurs sequentially on each episode as before. The choice of the learning tasks is made in such a way that the learner is able to learn a \emph{concept} easier than if they were given the hardest task initially. Here, rather than attempting to learn a useful bias, we assume that knowledge of such a bias, and generate a sequence of task that respect this bias. We will not discuss this technique further.

% \begin{itemize}
% 	\item meta learning
% 	\item transfer learning
% 	\item multitask learning
% 	\item lifelong learning
% 	\item continual learning
% 	\item online learning
% 	\item curriculum learning
% 	\item few-shot learning / zero shot learning
% \end{itemize}
\section{Existing Methods Implicitly use Foliated Structures} \label{sec:reformalising_exising_work}

In this section, we will provide examples of well established models that perform transfer, and how they can be derived from our framework. There are some points that are applicable to all the present cases. 

Firstly, all these models assume that given a set of tasks, we implicitly assume that all the tasks lie in the same set of related tasks. That is, we assume that they all exist on the same leaf of a foliation on the task space. In that sense, when trying to learn on the model space, the problem becomes one of finding the appropriate leaf given a chosen foliation, and the points along each leaf that can represent the given tasks. Thus the equivariance requirement of the learning algorithm is implicitly satisfied by the construction of the model space and the implicitly chosen foliation. 

Secondly, in these models, we are only interested in a local neighbourhood of the model space; it is likely that these regions also contain non-identified models, for example if a $\mathrm{ReLu}$ non-linearity is used. In fact, unless we are interested in global properties, it is sufficient to look at a local neighbourhood. 

The general strategy that we use here is to specify how each of these models create a split of the total degrees of freedom that the model space has into global and local parameters; the foliated structure is really just a way to generalise this idea. 

\subsection{Transfer Learning with Global and Local Parameters}
These types of models are commonly seen in the transfer learning setting \citep{Pan2009}, and in some meta learning models, such as \citep{Steindor2018, Janith2019, Hausman2018, Kaddour2020}. These models simply take the direct definition of the local coordinates of a foliation, and use that as the definition of the model. Figure \ref{fig:transfer_learning} shows an example network topology of such a model. 

In inductive transfer learning, a common approach is to pretrain a model on a large dataset, to obtain optimal parameters. Suppose these are $\theta$. Then, given a new dataset, the model is rectified into $(\theta_{fixed}, \theta_{retrain})$; the $\theta_{retrain}$ components are retrained while the $\theta_{fixed}$ components are kept fixed. We see immediately that this is simply describing the local coordinates of a foliation. 

That is, in a local open set $U \subset \model$, we have a chart $\phi$ such that $\phi(m) = (\theta_{fixed}, \theta_{retrain})$, where $\theta_{fixed} \in \R^a$, $\theta_{retrain} \in \R^b$ and $a + b = n$, the total number of parameters. Each $\theta_{fixed}$ defines a local plaque of dimension $b$. During transfer, since $\theta_{fixed}$ remains constant, the retraining occurs on this plaque.  

Something similar occurs in \citep{Steindor2018}. Here, the split between the fixed and retrained parameters is done at the same time. In this case, they are called the global and local parameters respectively. They are called as such because the global parameters apply to all tasks, whereas the local parameters are task specific. Furthermore, the global parameters are the data points themselves (since the model they use is a nonparametric Gaussian Process). These values are fixed, and the optimisation that is carried out is over the hyperparameters of the GP model, and a set of latent variables. These latent variables are written for each task. The hyperparameters could be thought of as a part of the global parameters, and the latent variables are the task specific parameters. Thus the learning objective finds a portion of the global parameters, as well as suitable values of the latent variables. During inference time, the hyperparameters are fixed, and a new  latent variable is learned.

\subsection{Prototypical Networks: Simple Meta-Learning}
Prototypical Networks are used for few-shot classification \citep{Snell2017}. It learns an emdedding function which can be used to compute a distribution over the different classes of the problem. The embedding function is used to compute a prototype vector in the embedding space for each class. A distribution over the classes for a new query point is then computed by running a softmax over the distance from the prototypes to the embedding of the new data point. This algorithm then carries out few-shot learning by generating the prototypes by applying the embedding function on the dataset given for the new task, and immediately computing the predictions on queries. During training, each iteration involves choosing a subset of the classes and the given data for each, and optimising the embedding function so as to minimise the negative log likelihood of the true predictions. 
\begin{figure}[t] 
    \centering
    \includegraphics[width=0.8\textwidth]{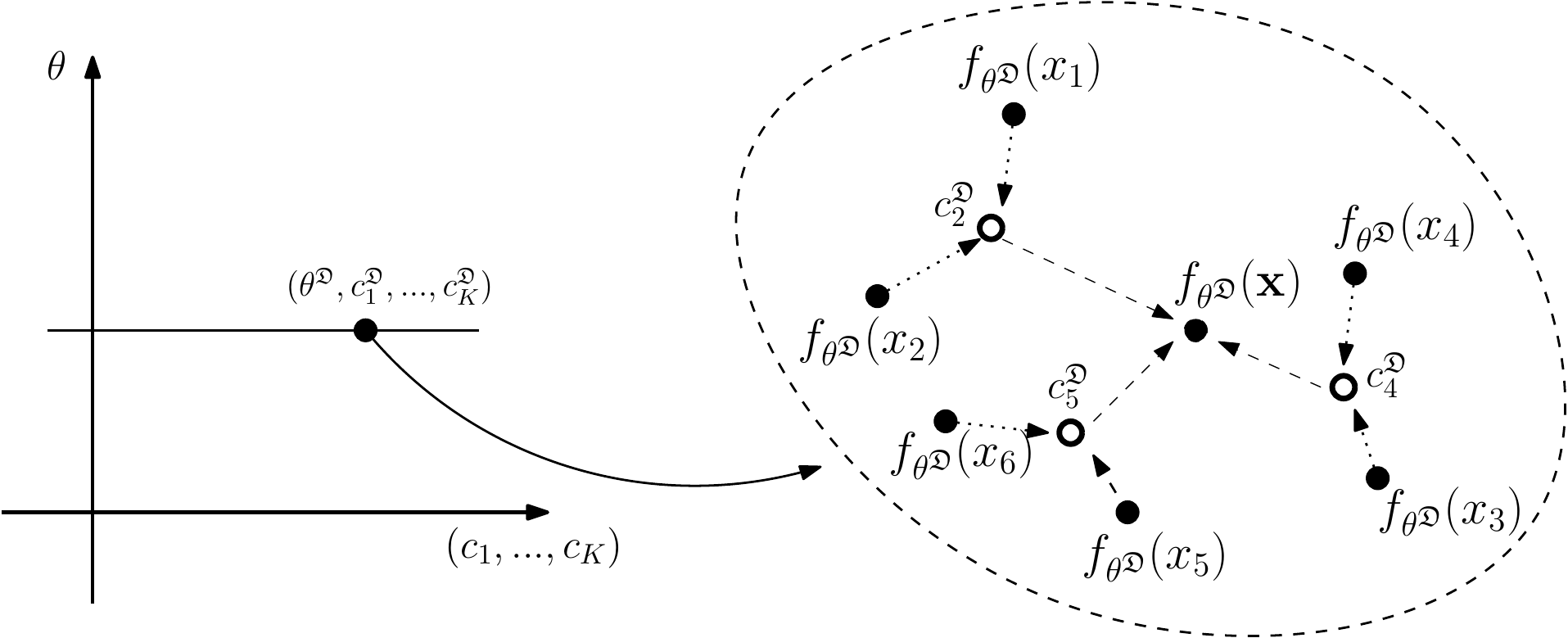}
    \caption{A foliation in the space $(\theta, c_1, ..., c_k, ..., c_K)$ gives us what we expect from a prototypical network. For a given dataset $\dataset$, we find $\theta^\dataset$ during the initial training phase. For a given task, which has classes from $\{2, 4, 5 \} \subset \{1, ..., K\}$, the values $(c_2, c_4, c_5)$ define the distribution over the class probabilities of a particular input $x$. These values are calculated from a small task specific dataset $\dataset_t = \{(x_1, 2), (x_2, 2), (x_3, 4), (x_4, 4), (x_5, 5), (x_6, 5)\}$, in this example.}

    \label{fig:prototypical}
\end{figure}
To be more precise, suppose we are given a data $\dataset = \{(x_i, y_i)\}_{i \in \index}$, where $x_i \in X$ and $y_i \in \{1, ..., K\}$ is a class identity. A classification task $t$ is given by choosing $n_t \leq K$ classes from  $\{1, ..., K\}$, denoted as $Y_t$ and $S \leq \|\index\|$ to generate a dataset $\dataset_t = \bigcup_{k \in Y_t}\{(x_i, y_i): (x_i, y_i) \in \dataset,  y_i \in Y_t\}$. The embedding function is a map $f_{\theta^\dataset}: X \rightarrow \R^m$, where ${\theta^\dataset} \in \theta$ are the parameters of this function, expressed as an element from a parameter space $\theta$. For the $k$-th class in $Y_t$, the prototype is given by,
\begin{equation}
c_k = \frac{1}{\|S_k\|}\sum_{(x_i, y_i) \in S_k}f_{\theta^\dataset}(x)(x_i).
\end{equation}
$S_k$ is the number of data points given for the class $k \in Y_t$. The probability of a particular class, given a distance metric $\rho$ on the embedding space, is then:
\begin{equation}
p(y=k\|x) = \frac{\exp(-\rho(f_{\theta^\dataset}(x), c_k))}{\sum_{k' \in \Y_t}\exp(-\rho(f_{\theta^\dataset}(x), c_{k'}))}.
\end{equation}

In this model, the full parameters that specify a prediction are $(\theta, c_1, ..., c_k, ..., c_K)$. $\theta^\dataset$ corresponds to the global bias that specifies how each input should be transformed; this is independent of the task that is currently being solved. The coordinates given by the remaining $(c_1, ..., c_k, ..., c_K)$ specify a distribution over the particular classes; they are the task specific parameters. Each $c_k$ therefore corresponds to a particular class. Since for a given task, $n_t \leq K$, we would be ignoring the $c_k$ coordinates that correspond to classes in the set $\{1, ..., K\} \backslash Y_t$. It should be noted that here, the adaption to the new task is deterministic and immediate; there is no \emph{training} (optimisation) required. This approach is illustrated in Figure \ref{fig:prototypical}.

\subsection{Model Agnostic Meta Learning (MAML)}
\begin{figure}[t] 
    \centering
    \includegraphics[width=0.4\textwidth]{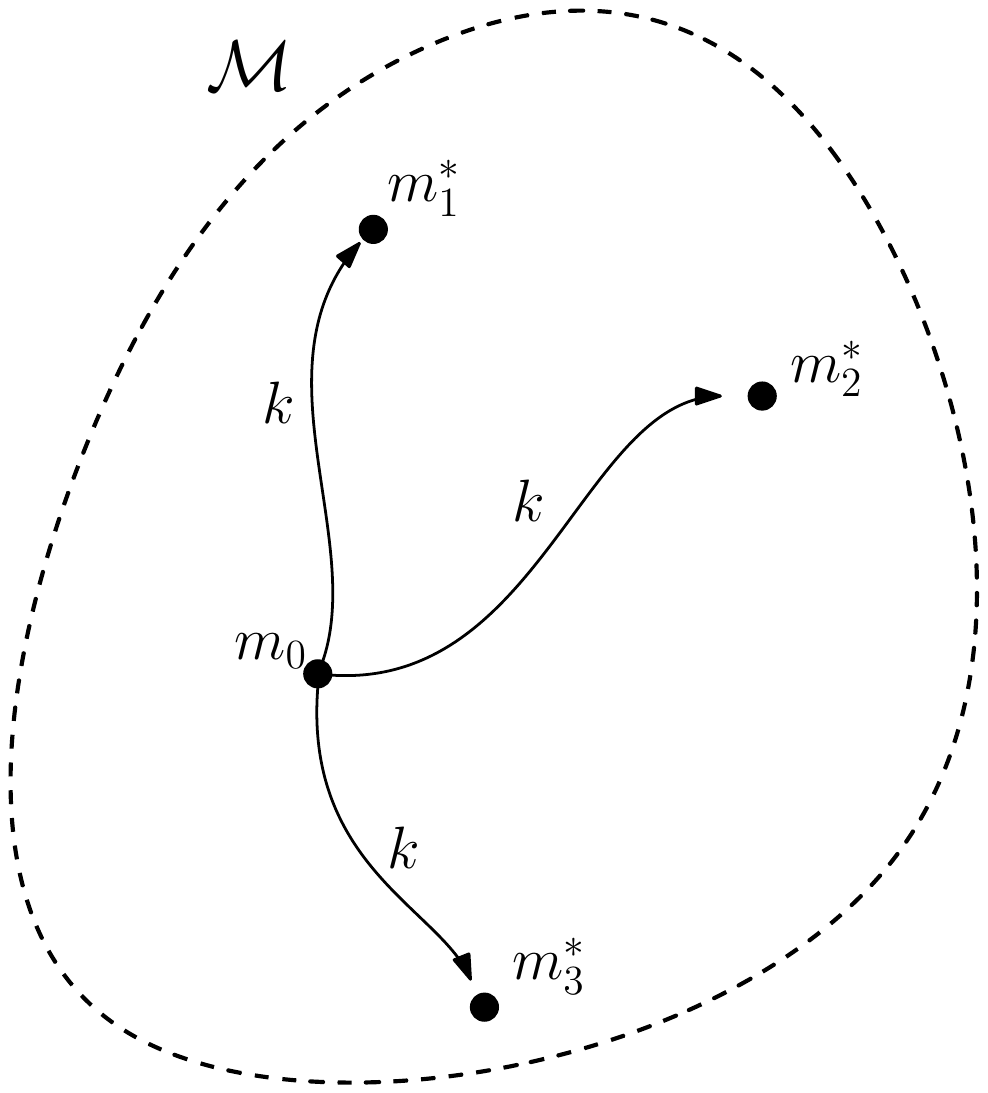}
    \caption{MAML finds an optimal initial model $m_0$ such that solutions to other tasks $m^*_1$, $m^*_2$ and $m^*_3$ are $k$ gradient descent steps away. }

    \label{fig:maml_standard}
\end{figure}

MAML \citep{Finn2017} is a meta-learning technique that has been very popular since it was derived a few years ago; it is a good example of a method that makes use of 2 levels of optimisation. For MAML, we are given a set of tasks; its goal is to find a common initialisation point, from where following gradient descent (or another gradient based optimisation method) for a fixed number of iterations will yield a suitable model for each task. MAML does this by optimising the initial point in a higher level loop, and following gradient descent for a fixed number of steps starting from this point; see Figure \ref{fig:maml_standard}. 

Showing that this type of model uses the structures we have discussed in general is difficult. This is particularly because we need to study the dynamics of the inner optimisation algorithm. This is made further difficult by the fact that in practice, MAML uses stochastic optimisers, and also only needs to reach a \emph{sufficiently} accurate model; while the latter point is also true in the previous cases, here, it can form a key part of the description of the model. We will make some simplifying assumptions regarding these; the dynamics of the descent algorithms follow the gradient flow of the vector field generated by the gradient of the loss function w.r.t the parameters of the model space. The task, per such descent, is fixed. 

If MAML is to carry out learning to transfer, then it must find some invariant structure that can be reused, without learning, when solving subsequent tasks. One can analyse MAML in several ways. One such method is given in \citet{Grant2018}, where it is posed as a hierarchical bayes model. Another method could be to look at it from a foliated point of view.  We make the following proposition:

\begin{figure}[t] 
    \centering
    \includegraphics[width=0.6\textwidth]{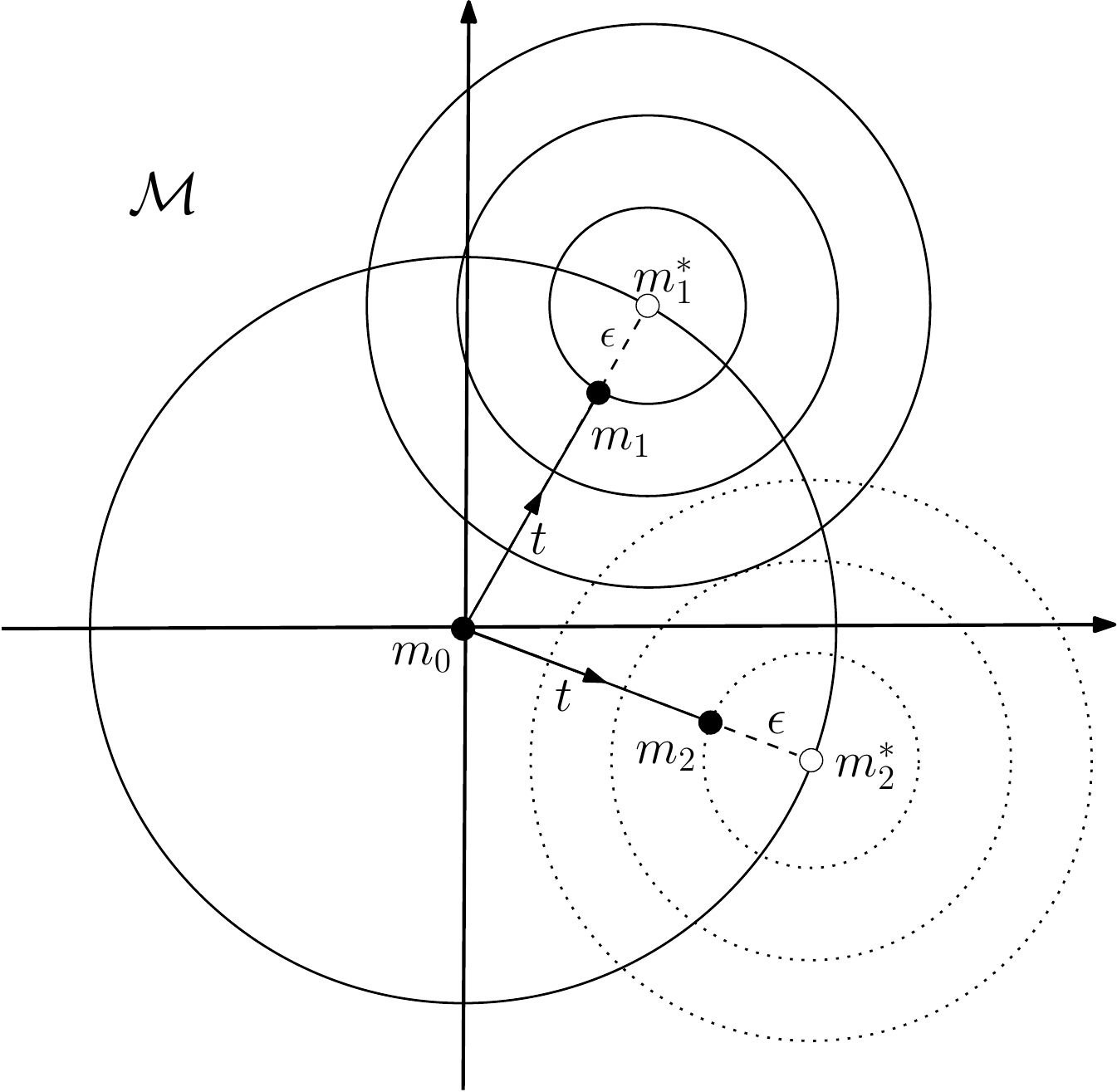}
    
    \caption{Showing the leaf generated by MAML for a loss function that is quadratic w.r.t the task space and the model space. Centered at each $m_{t_i}^*$, we have drawn some of the level sets of the loss function. This pictorially shows the claim that MAML finds points that are on a circle, when the coordinate system is centered at the starting point $m_0$. The arrows show the flows, starting at $m_0$ along the positive $t$ direction for each task.}

    \label{fig:maml}
\end{figure}

\begin{proposition}[Local foliation induced by MAML] \label{prop:maml}
We are given a task space $\task$, a model space $\model$ (a $d$-dimensional smooth manifold) and an \emph{appropriate} loss function $\loss: \task \times \model \rightarrow [0, \infty]$. Assume that $L$, restricted to some neighbourhood $U_\task$ of $\task$ and $U_\model $ of $\model$, is smooth and continuous w.r.t both variables. These neighbourhoods are chosen such that $\forall t \in U_\task$, there exists a unique  $m^*_t \in U_\model$, where $m^*_t = \argmin_{m \in U_\model}L(t,m)$.

Under certain conditions regarding how the gradient field of $\loss$ changes as the task is changed, and given a set of related tasks $R|_{U_\task}  = R \cap U_\task$ which is restricted to $U_\task$, we propose that MAML finds a point $m_0 \in U_\model$ such that for any task $t \in R|_{U_\task}$, we can obtain a solution $m_t$ where $\|\loss(t, m_t) - \loss(t, m_t^*)\| = \epsilon$, for $\epsilon$ that is fixed for all tasks in $U_\task$, by following the flow $\theta_{m_0}: \R \rightarrow \model$ of the gradient field of the negative loss for $k$ time steps, starting at $m_0$. That is $\theta_{m_0}(k)$ for any task in $R|_{U_\task}$ is a solution that is of the prescribed accuracy $\epsilon$.  

Then, the set $\{m \in U_\model\}$ such that the above condition is true for all for a fixed $k$ and $\epsilon$ forms a leaf of a folation; other leaves can be reached by changing $k$. Thus MAML describes a $d-1$ foliation (locally).

\end{proposition}

The proof of this very general statement is beyond the scope of the present work. It requires us to closely study how a given loss function changes as tasks are transformed; this needs careful specification of task spaces and its notion of relatedness. There are some points to note here. We have made the accuracy fixed; in practice, we are okay with models that are \emph{at worst} of a chosen accuracy; in the view of the foliations, we are saying that several, neighbouring leaves are good enough for our solution. The same can be said about stating that we want to reach a sufficiently suitable solution in at most $k$ steps. It is possible that this requirement ends up creating a partition of a tesselation; that is we have found a region of similarity. All this will be explored in future work.

In order to present an intuition as to why the above would be correct, we present a very simplified example that illustrates Proposition \ref{prop:maml}. Let us assume that the model space is locally 2-dimensional; that is $\phi(U_\model) \subset \R^2$, for some chart on the $\model$. Write the same for $U_\task$. Further, make the very restrictive assumption that the loss function $\loss: U_\task \times U_\model \rightarrow [0, \infty]$ is homogeneously quadratic in both variables. That is, we can write the loss as 

\begin{equation} \label{eq:loss_maml_simple}
\loss((t_1, t_2), (m_1, m_2)) = \sum_{i = 1}^2(m_i - t_i)^2
\end{equation}

Here, we see that changing the task merely changes the position of $m_t^*$, but keeps the shape of loss curve over $\model$ fixed. The level sets of this system create circles centered at $t$; this is shown in Figure \ref{fig:maml}. The loss value at the optimal model for a given model is $0$. For such as system, we can show the following theorem:

\begin{theorem} \label{theo:maml_simple}
    For a given $m_0$, and a coordinate system centered at $m_0$, the subsets of $\task$ for which the accuracy of the model is a fixed $\epsilon$, is given by 
    \begin{equation}
        \sum_{i=1}^2t_i^2 = \frac{\epsilon}{e^{-4k}}. 
    \end{equation}
\end{theorem}

Here, $(t_1, t_2)$, are the coordinates of each of the dimensions of a point $t \in \task$. $k$ denotes a variable that controls the radius; this $k$ corresponds to time that one must follow the flow of a task starting at $m_0$ to end up at the model that is of $\epsilon$ accuracy. The following corollary can then be shown: 

\begin{corollary} \label{coro:maml_simple}
    Given a task $t = (t_1, t_2) \in \task$, a chosen accuracy $\epsilon \in [0, \infty]$, and an initial model $m_0$, the time $k$ needed to get to a model $m_t$ for which the accuracy $\|\loss(t, m_t) - L(t, m_t^*)\| = \epsilon$ is given by: 
    \begin{equation}
        k_\epsilon = 4\ln\bigg(\frac{\epsilon}{t_1^2 + t_2^2}\bigg),
    \end{equation}
    and the model coordinates $m_t = (m_1, m_2)$ are given by: 
    \begin{equation}
        m_i = -t_i\bigg(\frac{\epsilon}{t_1^2 + t_2^2}\bigg) + t_i.
    \end{equation}
\end{corollary}
The proofs for Theorem \ref{theo:maml_simple} and Corollary \ref{coro:maml_simple} are given in Appendix \ref{sec:maml_proof}. These equations show that the subsets that satisfy the criterion for a set of related tasks under the MAML scheme are \emph{circles} in $\R^2$, for the given problem specification; this is a regular foliation on $\R^2 / {0}$. These results can be extended to higher dimensions, and to loss functions that  aren't symmetric; for example, if the level sets look like ellipsoids, then we can expect that leaves would be ellipsoids too.

In this example, the behaviour of the loss function was specified from the outset. Typically, under the targeted equivariance of the learning algorithm, this behaviour would be derived from the defined charts of the task space, the assumed notions of relatedness on the task space, and the chosen foliation (and therefore charts) on the model space. In Equation \ref{eq:loss_maml_simple} (and the subsequent proofs) for example, we have assumed that the task space has trivial coordinates in $\R^2$. It therefore stands to reason for the existence of a class of loss functions, for which the conditions of the MAML specification are satisfied. That is:

\begin{proposition} [Loss functions for MAML] \label{prop:maml_loss}
    We are given a task space $\task$ and a model space $\model$ which are described as smooth manifolds with foliations. $U_\task$ and $U_\model$ are open subsets of $\task$ and $\model$ respectively. Given a notion of relatedness, there is a local foliation on $U_\task$, and an associated local foliation on $U_\model$ such that a leaf on $U_\task$, denoted $\leaf_{\task}$ has a corresponding leaf on $U_\model$, denoted as $\leaf_{\model, \task}$, which contains \emph{the best} or \emph{sufficiently good} descriptions of the elements of the former. For a given $t \in U_\task$, $m_t \in U_\model$ is this descriptor.

    Any loss function $\loss_t$ on $U_\model$, restricted to $t \in U_\task$ will produce a flow on $U_\task$. Suppose we choose an $m_0 \in U_\model$ and a $k \in [0, \infty]$. We propose that there exists a set of loss functions $\loss_{\mathrm{MAML}}$ such that for a task $t \in U_\task$, $\theta(k, m_0)_{\leaf_t} = m_t$, and that if $t \in \leaf_\task$ for some leaf, then for all such $t$, $m_t \in \leaf_{\model, \task}$.
\end{proposition}

This proposition can be rephrased in terms of the transformations that form the leaves on the task and model spaces. That is, if $\pi_\task$ and $\pi_\model$ are such transformations, then for $s = \pi_{\task,ts}(t)$, and $m_s = \pi_{\model,ts}(m_t)$,
$\theta(k, m_0)_{L_s} = \pi_{\model,ts}(\theta(k, m_0)_{L_t})$.

Proposition \ref{prop:maml_loss} tells us that there are several components that can be controlled to gain the learning system that we want. In this case, we have chosen $k$, $m_0$ and the notions of relatedness, and are trying to construct a loss function appropriately. In the original MAML problem, we are trying to find the $m_0$, given the rest of the components. Proving this proposition is beyond the scope of the present paper. 

It must be noted that the condition where $\|\loss(t, m_t) - \loss(t, m_t^*)\| = \epsilon$, stated in Proposition \ref{prop:maml} might make it difficult to prove the existence of the proposed foliation. In Theorem \ref{theo:maml_simple}, we showed that it can exist in a simple case; in particular this depended heavily on how the gradient fields of the loss functions changed as we changed that tasks. As such, an alternative method of looking at MAML is to consider when $\|\loss(t, m_t) - \loss(t, m_t^*)\| < \epsilon$ is allowed. In this case, it is possible that MAML describes an open set around $m_0$ which corresponds to a set of related tasks. It doesn't define a global foliation, or any global structure at all. In both cases, MAML parameterises a single set of related tasks, and moves this set until the observed tasks fall within it. Further analysis of such a picture of MAML is left for future work.

\section{Discussion} \label{sec:discussion}
In this section, we will provide a discussion of some key aspects of the framework we have presented.

\subsection{Examples of Relatedness} \label{sec:examples_of_relatedness}
In Section \ref{sec:relatedness}, we provided a definition of a notion of relatedness in terms of groups and/or pseudogroups. In this section, we will provide some examples of this, to help illustrate the concept.

\begin{figure}[ht] 
    \centering
    \includegraphics[width=0.8\textwidth, height=3in]{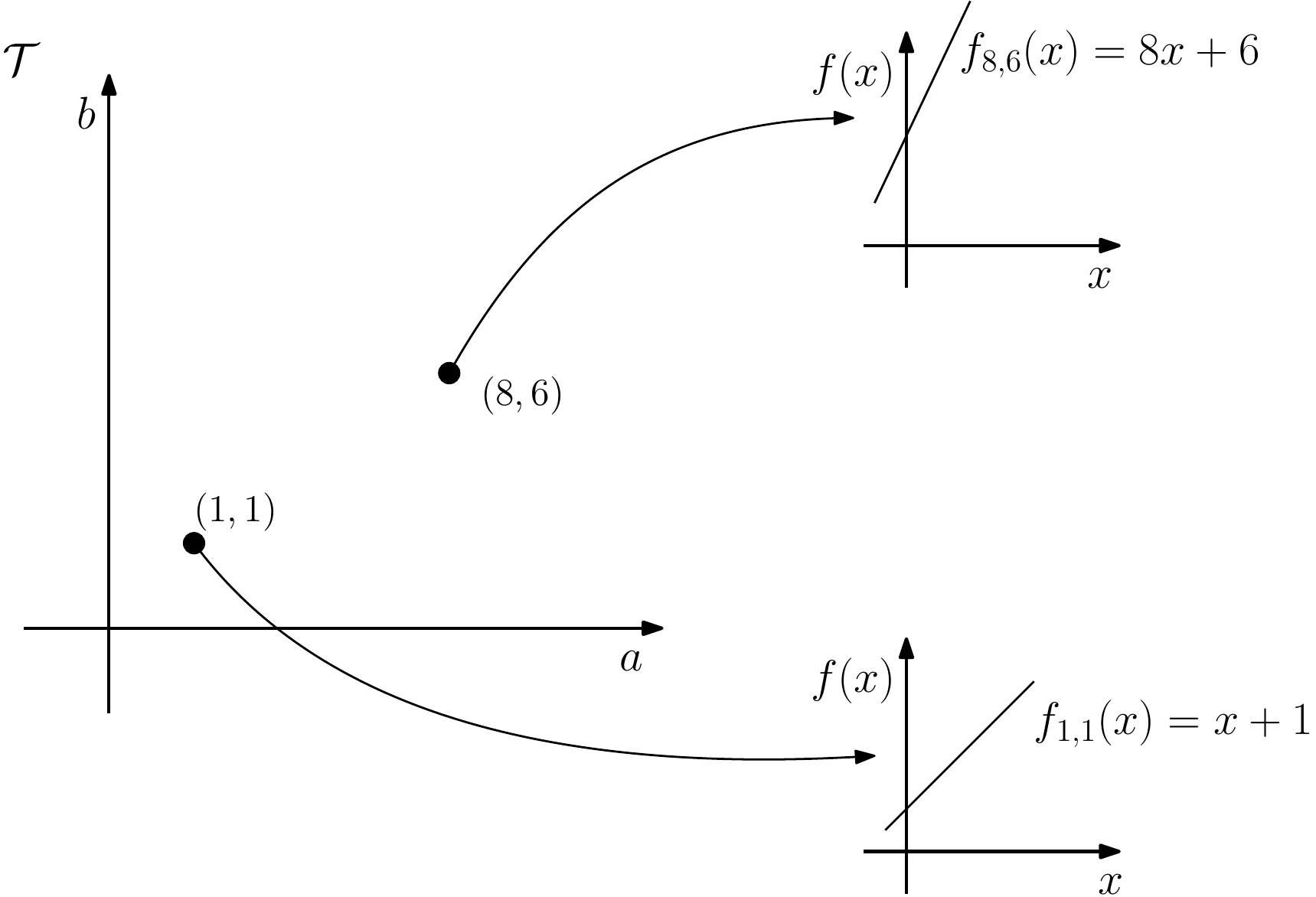}
    
    \caption{A simple task space $\task$, where its structure is defined by functions $f_{a,b}(x) = ax + b$, for $a,b,x \in \R$.}
    \label{fig:simple_task_space}
\end{figure}

Let us look at a simple problem. Consider a set of functions that can be written as $\task = \{f_{a,b}: \R \rightarrow \R \| a, b \in \R \}$, where $f_{a,b}(x) = af(x) + b$ for $x \in \R$, and $f(x)$ is some smooth and continuous function $f: \R \rightarrow \R$. Figure \ref{fig:simple_task_space} shows an example of this, where $f(x) = x$. We are interested in understanding a notion of relatedness between tasks in the task space. We can see that $\task$ is isomorphic to $\R^2$, where each point represents a particular function, with coordinates given by $(a, b)$. 

Imagine that from this set, we are given some learning tasks to solve that satisfy the constraint that $b=1$. For example, this set of learning tasks contain $t_1, t_2 \in \T_{b=1} = \{f_{a, b=1} \}$, with $t_1 = f_{a=1, b=1}$ and $t_2 = f_{a=2, b=1}$. The set $\T_{b=1}$ is simply a straight line, parallel to the horizontal axis in $\R^2$ space, and \emph{any element} of this subset can be accessed by choosing the appropriate value of $a$. The difference between the two tasks here can be represented in terms of the difference between their values of $a$. That is, we can transform from $t_1$ to $t_2$ by \emph{adding} $1$ to the first coordinate. The additive group $\R$ that acts on the whole of $\task$ is the notion of relatedness here; the horizontal parallel lines in $\R^2$ are the sets of related tasks.

A notion of relatedness provides us with standard by which to measure relatedness against. For example, suppose in a different problem, we were given tasks from a set $\T_{circ} = \{f_{a, b} | a, b \in \R, a^2 + b^2 = 1 \}$. In this case, we can see that $\T_{circ}$ can be represented by a unit circle in $\R^2$. The notion of relatedness then is the circle group $\Circ^1$.

\begin{figure}[ht] 
    \centering
    \includegraphics[width=0.8\textwidth, height=3in]{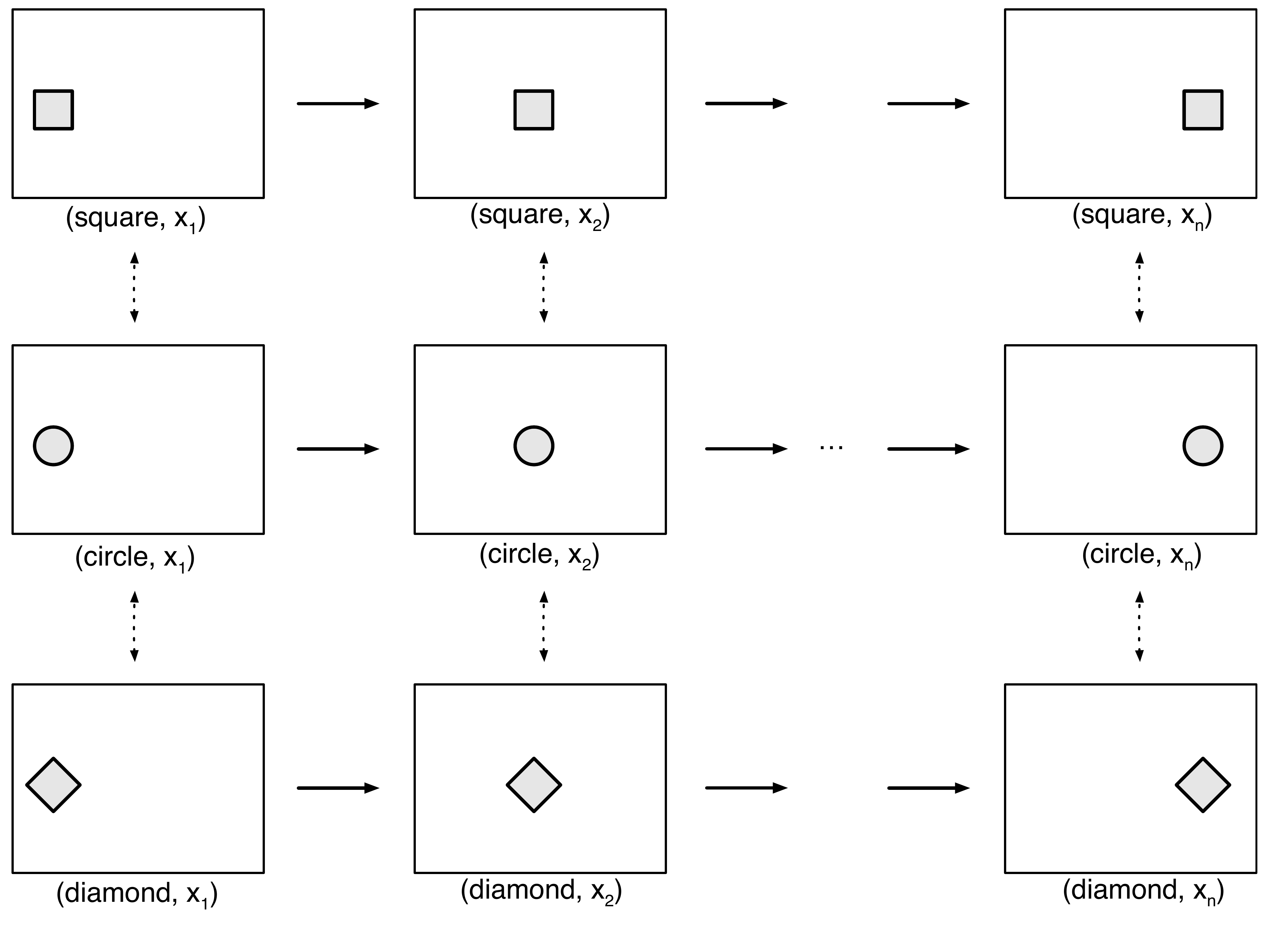}
    
    \caption{Images for image classification that contain some basic shapes. These shapes can be generated as the interior points of a set bounded by constant $L_p$ norms about origins given by $(x, 1)$; $p \in [0, \infty]$. The reference coordinate system has $(0, 0)$ at the bottom left corner of the image boundary. This set consists of all images obtained by horizontally and continuously translating the ``object'' in the image between the left-most and right-most columns.}
    \label{fig:image_translation}
\end{figure}

As a more complicated example, consider the diagram given in Figure \ref{fig:image_translation}. We are introducing a set of images that can be used for a classification problem; let us denote this set by $I$. There are several classification problems we can consider here. Let's consider the task of multi-class classification, where we must identify whether the object on the image is a square, circle or a diamond, and the objects are centered at $x_1$; thus we are looking at the first column of Figure \ref{fig:image_translation}. Imagine then that our dataset is translated rightwards, by moving the objects by some amount to the right; the classes still stay the same, and the task is to learn a classifier that will adapt to the transformed domain. Here, the transformations can again be thought of in terms of position along a circle $\Circ$; like in a game of Snake, once the object hits the right edge, it resets to the left most edge\footnote{If there was vertical movement, this group would be the torus $\Circ \times \Circ$.}. 

Note that this is qualitatively different from the regression task from the previous example. That is, instead of changing the qualitative nature of the structure, we have changed the domain. That is, we are still trying to map images of diamonds, circles and squares to their labels; instead the images are slightly altered. This is called domain adaptation in the literature \citep{Pan2010, Daume2006, Kouw2018, Ben2007}. As far as our definition of a learning task is concerned, there is no real difference, abstractly speaking. The map from input to output still has to change. 

Another problem setting we can consider from this domain is that of binary classifiers that make true or false statements about the identity of the image. That is, $f_{p, x}: I \rightarrow \{0, 1\}$, where $f_{p, x}(i) = 1$ iff the image $i$ consists of the shape given by the interior of the boundary made of points $t$ in $\R^2$ such that $||t||_p = 1$, and is centered at $(x, 1)$ in the reference coordinate system. In this case, we can see that an extra dimension has been added to the problems; this full space consists of the direct product $[0, \infty] \times \mathbb{S}^1$. We can then consider smaller subsets of related tasks to consider for learning to transfer; for example, each of the rows in Figure \ref{fig:image_translation} are circles of fixed $p$ in $[0, \infty] \times \mathbb{S}^1$.

In terms of the pendulum example, we can note that the mass and length values can be transformed as the direct product $[0, \infty] \times [0, \infty]$. The full task space that contains the pendulums can be the set of classical dynamical systems that vary with respect to a notion of a single length and mass. This space could include cartpoles, double pendulums where only a single pendulum changes between tasks, etc. 

% It is important to note that in these examples, the sets of related tasks are subsets of the task spaces. In the way that we have described things, we can clearly see that there is a particular set of transformations that is transitive to the task space. That is, there is a transformation set with respect to which the full task space is the set of related tasks; all tasks can be transformed to another. Such a transformation set is not useful to us. The idea with transfer is that we want to learn something about a set of tasks that makes learning a new, unseen task in that set easier. This something is the relationship between the tasks; once we know how tasks can be transformed from one to another, we only need to learn the transformation from the set of transformations. As we said before, this set is smaller, and so makes the learning problem easier. This relationship is the useful bias that we want to learn. 

\subsection{Invariances} \label{sec:invariances}
In the first example in Section \ref{sec:examples_of_relatedness}, we saw that we could create a set of related tasks by keeping $b$ fixed at $1$. Similarly, in $f_{circ}$, we had the radius of the circle at $1$, and in $I$, we could fix $p$. These show another, and possibly complementary characteristic of a set of related tasks; in addition to containing a notion of relatedness, there is also an \emph{invariant quantity} that stays constant within this set. The invariant quantity is derived from the transformation set. 

An invariant quantity is defined in the following way. 

\begin{definition}[Invariant quantity]
	Given a space $X$, a quantity on $X$ is a map $f: X \rightarrow \R$. This quantity is said to be invariant with respect to a transformation $\pi$ on $X$, where $\pi: X \rightarrow X$, iff $\forall x \in X$, $f(x) = f(\pi(x))$. 	
\end{definition}

This definition implies that irrespective of a particular transformation applied on the input space, the quantity $f(x)$ remains unchanged for any value of $x$. $g$ is the transformation with respect to which the quantity is invariant; it is a key component. In the definition, $g$ is a single function. It is often useful to define invariant quantities are described with respect to a set of functions $\mathcal{G}$, where for any $g \in \G$, the above condition holds. If the set of functions respects the definition of a group, then $\G$ is said to form a symmetry group of $X$. 

Such a group falls within our intuitive notion of symmetries. We know that a circle in $\R^2$ is symmetric under rotations; an invariant quantity here is the radius of the circle. That is, if we were to express the circle in Cartesian coordinates, then the radius $r(x, y) = \sqrt{x^2 + y^2}$ does not change if were to rotate about the centre by some angle. Now the transformations do change the superficial ``quantifiable value'' of a point. It does not however change the ``nature'' of it; the nature of the object then, is given by its invariant quantities. This is equivalent to saying that the nature of the object is defined by being symmetric with respect to the given group. That is, considering a set of ``all possible things'', an invariance can be used to collect a subset of this set such that they have a consistent and constant property.

The notion of symmetries, and invariances can therefore be used to identify system. In physics for example, particular invariances are a characteristic feature of a physical system \citep{Wigner1949}, in that there are some canonical groups which our laws of physics are symmetric to. In classical systems, we expect  the behaviour of systems to be invariant to translations in space and time. For example, the way that a pendulum swings (assuming a closed system with a uniform gravitational field) should not depend on \emph{where} in the laboratory the experiment is carried out. It also should not depend on whether the experiment is started at midnight, or at noon. Thus, from all possible ways that a physical object should behave, the assumed invariances focus our search to a smaller set of behaviours. 

There are of course other invariances that are needed to uniquely define the behaviour of a system. In the pendulum, these include the fact that the motion of a pendulum occurs around a circle, since the length of the pole is fixed. In a double pendulum, another such constraint\footnote{Note that we have not distinguished between constraint or invariance. A constraint is often described as a function $f_{con}(x) = 0, \: \forall x\in X$. This looks remarkably similar to the definition of an invariance; a constraint implicitly defines an invariance.} is that the 2 poles are connected. A set of invariances then allow us to identify the particular system we want from the set of all systems, in this case classical systems. When we apply Newton's Second Law to a physical problem, we explicitly write down these invariances. Furthermore, classical systems themselves obey  Hamilton's principle where the path taken over time through its generalised coordinates is an extremal of its action functional \citep{Arnol2013}; this distinguishes this set from the set of all possible physical systems. 

The pendulum is also commonly used in RL problems, as we described in Figure \ref{fig:pendulum_example}. Changing the mass and length of the pendulum will also change the policies in a similar way. The invariant quantity in the set of RL tasks must also contain details about the control problem at hand; that is the fact that we always start at the bottom, and want to swing up and balance at the top, in the fastest strategy possible. It will also contain any other control constraints that are enforced in the problem. We could consider the case when the goals of the problem change, as was done in \citep{Janith2019} in the case of the cart-pole dynamical system. 

\begin{figure}[t]
    \centering
    \includegraphics[width=1\textwidth]{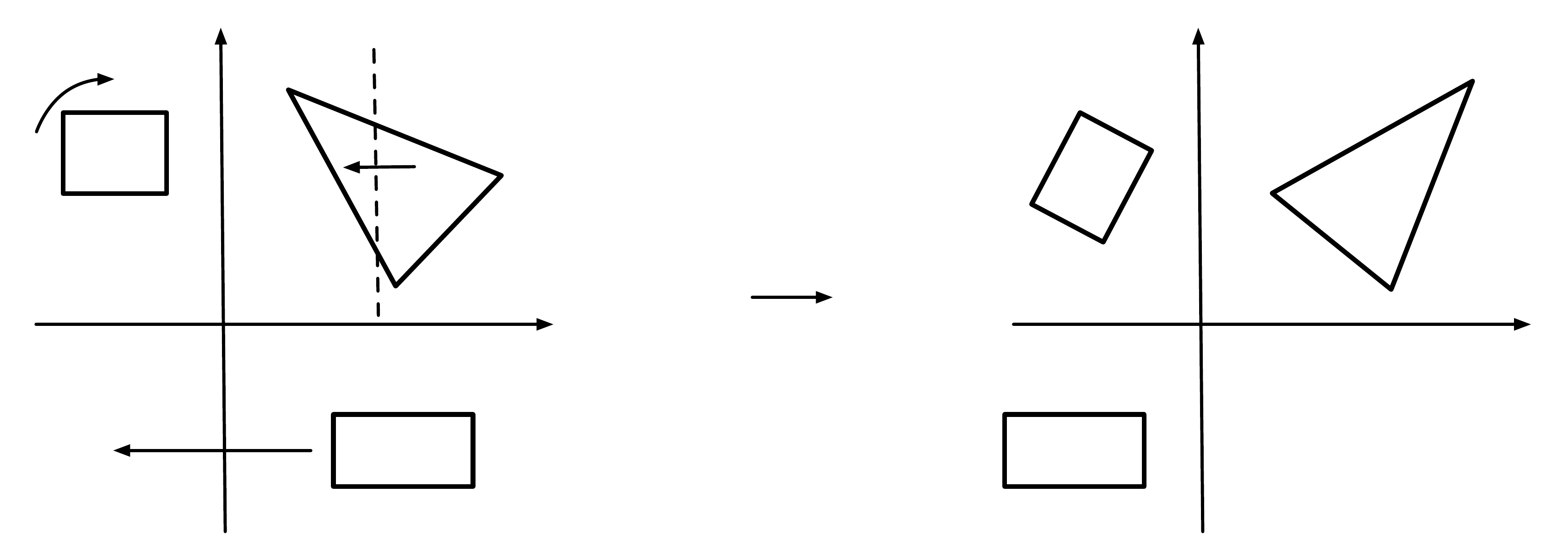}
    
    \caption{Rigid body transformations in $\R^2$ on a squares and triangles. These transformations include rotations, reflections and translations.}
    \label{fig:invariance_rotation}
\end{figure} 

An invariant quantity, along with its corresponding transformation, can be used to identify a class of objects from within a larger set. Consider the example in Figure \ref{fig:invariance_rotation}. Here, we have 2 objects, a square and a triangle. The transformation group we are considering is the union of reflections, rotations and translations in $\R^2$; together, these form the rigid body transformations. We will denote this set as $\G_{RB}$. 

Suppose we considered the quantity which is defined as the area enclosed by each object. The area is a map $A: ST \rightarrow [0, \infty]$, where $ST$ is the set of all possible triangles and squares, of different sizes and orientations. Another quantity that can be consider is the number of vertices; $N: ST \rightarrow \N$. It is clear that under a transformation from  $\G_{RB}$, the area of a particular shape does not change. This is true whether we are talking about a square or triangle. This is true for any polygon or closed curve in $\R^2$. We can therefore say that, a defining characteristic of closed curves in $\R^2$ is that their areas are invariant under rotations. The defining characteristic of $ST$ is that it \emph{also} only consists of polygons with 3 or 4 vertices; of course, $N_{ST}$ is also an invariant quantity here. 

In our framework, where notions of relatedness are represented in the learning problem as a foliation, there is a natural, local set of invariant quantities that arise. This is most clear on a regular foliation, where we know that a subset of the local coordinates of a chart remain constant. 

\subsection{Similarity} \label{sec:similarity}
A related concept that is often used to compare tasks is \emph{similarity}. Similarity is often used interchangeably with relatedness; here we will present a distinction. 
\begin{definition}[A Set of Similar Tasks]
    A metric $\rho$ on $\task$ is called a notion of similarity. $t_1, t_2 \in \task$ are said to be similar if $\rho(t_1, t_2) < \epsilon$. Then, given $t_1 \in \task$, the set $\{t \in \task | \rho(t, t_1) < \epsilon \in \R^+\} $ is called a set of tasks similar to $t_1$. 
\end{definition}
If we choose a countable number of reference tasks on $\task$, then a notion of similarity can induce a partition on $\task$ in the way of a Voronoi diagram \citep{Reddy2012}. That is, we can write an equivalence relation, given a set of reference tasks $R_\task = {r_i \in \task}_{i \in \index}$, $t_1 \sim t_2 \iff \argmin_{r_i \in R_\task}\rho(r_i, t_1) = \argmin_{r_i \in R_\task}\rho(r_i, t_1)$. Such a partition would look like the tesselation in Figure \ref{fig:tessellation}. Since this is a partition, there is a notion of relatedness that can be associated with it (for example, by trivially considering the permutation groups on each element of the partition). 

It is possible to consider transfer in terms of similarity, by transferring within a set of similar tasks, as defined above, given a task $t \in \task$. This is often carried out in Transfer Learning \citep{Pan2009}, or Continual Learning \citep{Lesort2020}, where pre-trained models are \emph{slightly} updated to a new dataset. Our notion of similarity can describe why catastrophic forgetting occurs \citep{Goodfellow2013} in such methods. The further away from the original pre-trained model we move, the less similar the new models are; thus, gradually, we lose any resemblance to it. In the case of transfer using relatedness, such an issue does not occur as long as we fix and stay on the parallel space we operate on.

\begin{figure} 
    \centering
    \includegraphics[width=0.4\textwidth]{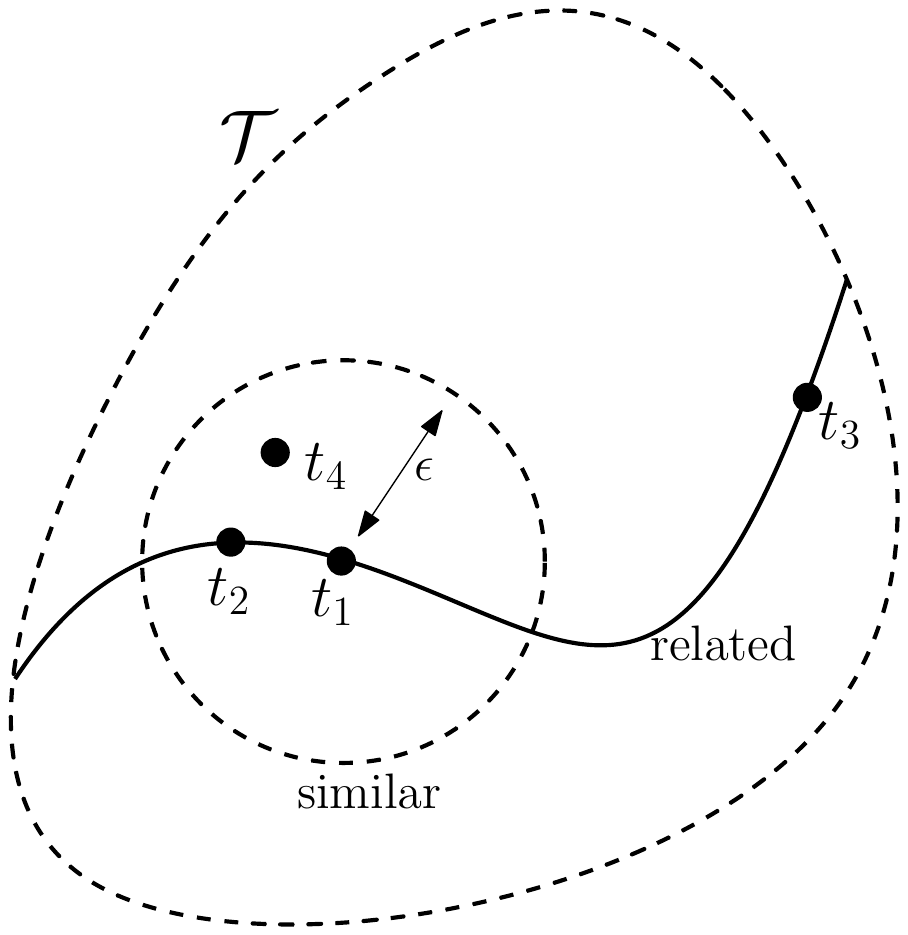}
    \caption{The distinction between similarity and relatedness. Similarity is defined in terms of an $\epsilon$-ball around a particular element. Here, points $t_2$ and $t_3$ are similar to $t_1$. On the other hand, relatedness is defined in terms of a transformation set; this relationship is shown as a line joining them. Thus, tasks $t_1, t_2$ and $t_3$ are related to each other. However, $t_4$ is not related to $t_2$ and $t_1$ though they are similar; $t_3$ is not similar to $t_2$ and $t_1$, though they are related.}
    \label{fig:related_vs_similar}
\end{figure}

There are some notable practical differences between relatedness and similarity. In the former the consistency of parallel spaces depends on the set of transformation groups or pseudogroups that are chosen. That is, if we were to pick out two tasks that are related to each other, then a third task, which is related to the second would also be related to the first, all with respect to the transformation set. In particular, since a notion of relatedness can give rise to a partition of $\task$, it provides us with some global structure that can be used for transfer.

On the other hand, a set of similar tasks is only useful locally (unless it is used to generate a tesselation, in which case there is an induced notion of relatedness), since relative to a given task, we can talk about a single set of similar tasks. In particular, the transitivity we mentioned previously, does not hold, unless the reference task is kept constant. Thus if $t_1$ being similar to $t_2$, which is similar to $t_3$ does not necessarily imply that $t_1$ is similar to $t_3$.

\section{Some Philosophical Considerations} \label{sec:philosophy}
In this section, we will describe some philosophical considerations of machine learning in general, and how our framework for learning to transfer carries naturally from them. 

\subsection{The Structure of Machine Learning} \label{sec:ml_structure}
In Definition \ref{def:learning_task}, we defined a learning task in terms of some structure that it contains, and uniquely identifies it. We hinted that this structure is often written as maps that carry certain properties. In this section we will describe such structure in more detail as applied to different problems in ML. The clearest example is supervised learning \citep{Bishop2006}. 

Supervised learning considers labelled data, which are generated from a joint distribution $p(X, Y)$ over input and output random variables $X$ and $Y$ respectively. We could consider that $p(X)$ is known, and since $p(X, Y) = p(Y)p(X | Y)$, we are really interested in learning the conditional distribution over $Y$ given $X$. An equivalent view point is to assume that there is a generative map $f: X \rightarrow Y$, which has some noise added to it; that is, $y = f(x) + \epsilon$, where $\epsilon \sim \mathcal{N}(0, 1)$. The map $f$, and its defining properties is the structure of the problem, and learning task is this entire structure. 

The learning task in RL often contains the policy. A RL problem is usually defined by a Markov Decision Process (MDP) $(S, A, T, r, \gamma)$; that is, given a system's transition dynamics $T = p(S'|S, A)$ which dictate the evolution of the system's state $S$, we want to find a policy $\pi = p(A|S)$ that maximises the long term return $R = \sum_{t=0}^{T}\gamma^{t+1}r(s_t)$ over the path taken by the state. $r$ is a reward function and $\gamma$ is a discount factor that describes how myopic the system is. The tuple $(S, A, T, r, \gamma)$ is the structure of an RL problem. Depending on the class of RL, the learning task is the subset of the structure that corresponds to the policy (model free), or $T$ and $r$ too (model based).

Unsupervised learning can be slightly more complicated. Let us consider the clustering problem \citep{Madhulatha2012} as an example. Here, we are given a set of unlabelled data from some domain $\X$, and we would like to identify clusters to which each data point belongs to. Suppose we assume that there exist an indexed set of $n$ clusters $\{c_i\}_{i=1}^n$ where each $c_i \in \X$, in unsupervised learning, we want to learn a map $f:  \X \rightarrow \{c_i\}$. In k-means \citep{Bishop2006}, this map outputs the cluster which is the shortest distance (based on some metric on $\X$ such as the Euclidean distance) to the new point. In kernel k-means \citep{Dhillon2004}, the point is first mapped onto a different space, before the same algorithm as k-means is run; this allows us to consider a different distance metric. The structure here are the clusters, and their members. 

More modern problems such as semi-supervised learning \citep{Zhu2009} can also become complicated when trying to define the precise structure we want to find; however, such a structure does exist. In semi-supervised classification for example, a dataset that consists of a mix of labelled and unlabelled data is given. The ratio is often skewed heavily towards the unlabelled data. The goal is to find a classifier using the total dataset that is \emph{better} than a classifier found using just the labelled data. This could, for example, be done by assuming that the data that map to a particular class clusters around some center (as in k-means). The classification aspect of this is supervised learning. The assumed structure is found in the assumption that the unlabelled data follow some regularity with the labelled data.

\subsection{Machine Learning as Learning Structure and Representations} \label{sec:representations}

Given a learning task and a model space, the problem of learning is to find an element of $\model$ that best \emph{represents} the learning task. This was described in Definition \ref{def:single_task_learning}. The choice of the word \emph{represent} here was no accident. This view of the learning problem was inspired by Marr's visionary book \emph{Vision} \citep{Marr1982}. A key component of this is the computational theory. 

%What is a computational theory
The computational theory of a system embodies \emph{what} the system must do. These are written in terms of transformations that must be carried out on the information that is received (if any), and the information that must be output. The simple example Marr presents is that of a cash register. The inputs it receives are the prices of items in a shopping cart; the output it must provide is the sum total of these prices. The action the cash register must do to accomplish this is addition. As simple as this sounds, it is important to understand that addition here is an abstract notion that carries with it some necessary rules. 

%Example of a computational theory for a cash register
The addition that the cash register must carry out is its \emph{process}; it is what it must do to its inputs to obtain its output. Now one can think of such a system that can carry out the process of addition abstractly, not just as a cash register. A generative process as we described in Definition \ref{def:learning_task} is such a process. In order to make the abstract real and useful, we need a \emph{representation} of the objects that are its inputs, outputs and its process. In \citep{Marr1982}, a representation is a formal \emph{scheme} by which an object can be described; the result of the application of such a scheme to a particular object is its \emph{description}. Thus,  the goal in ML is to find a description of the generative process w.r.t.\ a chosen representation. Often, some aspects of the process are known, or assumed; the remaining unknown structure is what we called the learning task. By defining model space, we have written a representation, and the learning problem is to find the best description of the learning task.

\begin{figure}[ht] 
    \centering
    \includegraphics[width=0.3\textwidth]{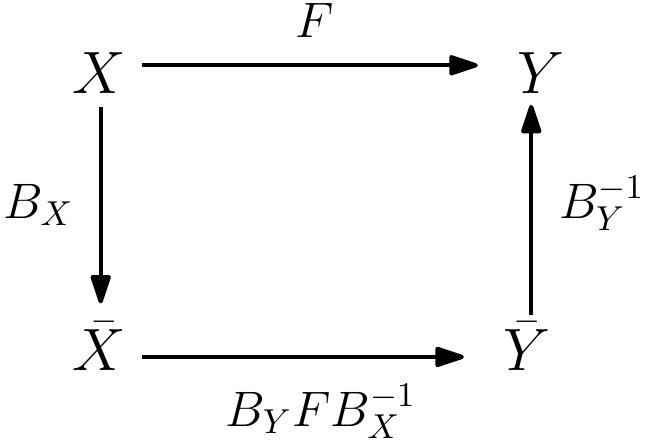}
    \caption{Commutation of linear maps under basis changes. }
    \label{fig:linear_map_commutation}
\end{figure}

As an example of a representation, we can look at number systems. The Arabic numeral scheme expresses integers as $(x \in \mathbb{N}) := \sum_ia_i10^i$, where $a_i \in \{0, 1, 2, ..., 9 \}$. Therefore, by writing the abstract number forty-two as ``42'', we are setting $a_0 = 2$, $a_1 = 4$ and the rest to zero. The same object $42$ can be similarly be described in the binary representation scheme as $101010$. The representation of addition operation then depends on the representation chosen of its inputs and outputs. Addition in the Arabic numeral system and the binary system are conceptually similar, since we add each unit, and carry over if the sum becomes larger than $9$ or $1$ respectively. If however, we had chosen a Roman numeral system, where $42$ is written as XLII, then addition will be carried out using different, more complicated rules, even though it is abstractly doing the same thing. 
%
%Representations are not unique
Representations satisfy some properties. A representation is not unique; there can be many ways to describe a single system. The important aspects of a representation that must be preserved between different representations are the behavioural properties that are important to the task of the system. In the case of a cash register, any representation of prices must be able to carry the additive group structure. It is then possible to consider mapping between suitable representations. Perhaps this is more conceivable if one considers vector spaces and linear maps between them. Suppose $f: X \rightarrow Y$ is a linear map that is written as a matrix $\mathbf{F}$ under the bases $e_X$ and $e_Y$ for $X$ and $Y$ respectively. If there is a basis change by transformations $\mathbf{B_X}$ and $\mathbf{B_Y}$, then under the new bases, the linear map will be described by $\bar{\mathbf{F}} = \mathbf{B_YFB_X^{-1}}$. $\bar{\mathbf{F}}$ and $\mathbf{F}$ are still the same linear map; it was merely their descriptions that changed when we changed the representation scheme; see Figure \ref{fig:linear_map_commutation} for the commutation graph for verification. Thus, all representations are equivalent up to isomorphisms. 
%
%Primitives build representations
Each representation can be thought to be composed of some elementary \emph{primitives}. These primitives are basic units that are combined, in some way, to describe all things a representation must describe. In the case of the Arabic numerals, these are the powers of 10, in binary, the powers of 2, and in vector spaces, the basis vectors. The representation also depends on \emph{how} these primitives are combined. In particular, we see that in the Arabic numerals, the powers of 10 are combined by a summing their multiples with one of $\{0, 1, ..., 9\}$. In binary, this set is $\{0, 1\}$, and in vector spaces, it is $\R$. As a more complicated example, consider an image, often represented as grid of pixels. In a grayscale image, each pixel can take a value from the closed set $[0,1]$, and the image cosists of the concatenation of several such numbers from $[0, 1]$ in a 2D grid. 

%Representations can approximate as well 
The example of an image brings forward another important aspect of a representation; it is possible for a representation to only \emph{approximate} the objects describes. An image (as we consider here) itself is a representation of some macroscopic scene. Depending on the camera that is used, such an image can be of different resolutions, where each pixel now describes some average of the region of the scene that it covers. However, the approximate representation must still satisfy whatever properties the exact representation does. Take the example of the cash register. A suitable approximating representation for prices are the positive integers, where each price is approximated by rounding up or down. Such a system still satisfies the requirement for the additive structure. That is, suppose the prices of 2 items are £5.4 and £7.10. In this approximate scheme, these prices are £$5\pm0.5$ and £$7\pm0.5$. The sum of the original prices is £12.50, which is £13 when approximated; the latter is within the accuracy of  $(5\pm0.5) +(7\pm0.5) = (12\pm1.0)$.

%Some representations are better
Finally, there is a notion by which some representation schemes are better than others; in the example of addition, as a process that humans can carry out mentally, the Arabic numeral scheme makes the arithmetic easier than if we were to use the binary or Roman numeral schemes. For use in computers however, the binary scheme is better. 

A model space we had defined previously can be used as a representation of $\process(\structure)$. A model in $\model$ is assumed to be able to act as a suitable description for the learning task at hand. Of course, as with Marr's representations, this element doesn't necessarily have to be \emph{exact}, just \emph{suitable}. Furthermore, there can be many model spaces that can similarly describe the task at hand, and as is fairly obvious with the most famous example of a parametric model, neural networks, these spaces can be built from elementary primitives. While each of these is a non-unique, suitable representation, they do however, contain other salient differences; these are their inductive biases.
 
\subsection{Inductive Bias and Learning to Transfer}
Each model and task space contains some inductive bias. Inductive biases are important for learning. Albeit not in familiar language, \cite{Mitchell1980} describes this eloquently. In a world where learning is described as finding a generalisation that is \emph{consistent} with a set of instances, a bias is anything that will make a particular generalisation, contained in the set of all possible and correct generalisations, more likely to be chosen. An unbiased generalizer assigns equal probability to each suitable generalisation. A bias introduces additional information or assumptions that have been made of the problem at hand, which make some equally suitable solutions (according to a purely data-fitting point of view) more relevant. 

The No Free Lunch Theorem \citep{Wolpert1996, Shwatz2014, Ho2002} is a more formal description of this idea that appeared a few years later. It states that no general purpose learning algorithm that can be usefully applied to \emph{any} problem is possible; there will always be a problem that another, more specialised or biased algorithm will outperform it in. 

This can also be illustrated in the bias-variance trade-off in classical learning theory \citep{Shwatz2014}. The error associated with a model of a hypothesis class can be decomposed into the sum of approximation and estimation errors. The approximation error is the lowest possible error in the hypothesis class; this of course independent of any data that is gathered. The estimation is the difference between this and the full error; this is therefore the error associated with choosing a hypothesis that is not the best in class, and depends on the data and the learning algorithm chosen. The approximation error can be reduced by increasing the complexity or the \emph{size} of the hypothesis class; this is reducing bias. This will invariably increase the estimation error since it is harder to search for the best element in a larger set; this is the variance.

More bias increases the approximation error, but reduces the estimation error. Classical theory, such as the VC dimension \citep{Vapnik2015} was devised to determine how to increase the bias such that the approximation error doesn't grow significantly faster than the estimation error. This will then reduce the overall error. 

Bias represents external information that is implicitly assumed to true about the learning task, and in particular, a set of related tasks. Thus bias creates subsets of \emph{all possible tasks}. Bias contains the subset of the structure of a learning problem that  is not \emph{learned} by the learning algorithm, but is simply assumed to be true. That is, it contains $\structure - t \subseteq \mathrm{bias} $, where $t$ is a learning task as defined in Definition \ref{def:learning_task}. When solving a single learning task, biases can define a representation scheme (or model space) from which to choose a description (or model) for the learning task. In particular, biases allow us to make a choice between competing, equally competent descriptions (from different model spaces). 

Humans for example, are biased towards simpler and more general models of things; see Newton's Laws for example. Simplicity has a certain \emph{intellectual romance} to it; Occam's razor could, for example, be posited as an imperative for elegance. Occam's razor requires us to chose the \emph{simplest} theory or model from a set of theories and models that can represent or explain the data observed equally well. The elegance arises from the lack of redundancies in the structure of a simpler theory. It is also the case that such a rule often works well in the kinds of problems that humans encounter. 

Such regularisation of model complexity can be found in ML methods. For example, choosing hyperparameter values of a Gaussian Process by maximising the marginal likelihood \citep{Rasmussen2006} explicitly penalises \emph{complex} models; such complexity is represented by one of the terms of the log marginal likelihood. Complexity here can be seen as the smoothness of the corresponding input-output maps of, say the posterior mean function; think the differences between the graph of a polynomial with low and high order. 

% Biases can also be introduced via domain knowledge from experts; past examples in similar fields or applications (as the expert determines) can guide the choice of how we write models. Many model architectures in ML presently are based on our experiences with the data that the model encounters; a good example of this are images, for which Convolutional Neural Networks (CNNs) were devised due to their spatial (translation) symmetry. 

% In addition, and in particular, biases will often depend on the application of the model that is being learned, the context in which the learning problem is written for. Occam's Razor is useful to us only because the types of problems that we encounter in science and philosophy contain useful solutions that follow this rule. This might not, in general be true; theories in string theory, quantum gravity and other outlandish ideas that have been devised to explain the physics of the universe are quite complicated. Of course one could argue that once the details have been hashed out and understood, a simpler theory would emerge; but even say Einstein's field equations for General Relativity, while elegant in principle, can be quite complicated once all the terms have been expanded. More relevantly, in ML, there could be a situation where a higher order polynomial like behaviour is more useful than a smoother, less complex solution. 

All this was to say that bias is important, and choosing the \emph{correct bias} is paramount to good generalisation. In learning to transfer, the notion of relatedness is a form of choosing a bias that is assumed to be applicable to each set of related tasks. That is, choosing a set of relatedness biases the model space in which we assume a suitable solutions to set of related tasks lie.

% This is true in single task learning, where the biases assumed The notions of transfer that we had defined are biases that can be placed on a space of tasks that generate different sets of related tasks

% In typical ML, this bias is often chosen a priori by experts, and controlled via regularisation terms in the loss function. The problem of transfer neatly fits in the idea of attempting to find a bias that is \emph{useful} for a set of tasks that is observed. The framework we described shows how that structure can be represented so that it can be learned principally. 

\subsection{Choosing a Notion of Relatedness} \label{sec:choosing_foliation}
As final thought, we bring attention to the problem of choosing the appropriate notion of relatedness. As mentioned, choosing a notion of relatedness is choosing a bias; the previous section argued that this is important. Such bias can also be introduced via domain knowledge from experts; past examples in similar fields or applications (as the expert determines) can guide the choice of how we write models. If it is known that the set of related tasks that are likely to be observed follow a particular notion of relatedness, then one can imbue their learning algorithm with the appropriate foliation. 

The more interesting case is when we can attempt to learn this structure. If our (task or model) space is given a foliated structure, we know that each leaf will typically be of a dimension less than that of the ambient space. A useful way of thinking about these dimensions is as factors of variations that allow us to reach all tasks in a set of related tasks. The search for a notion of relatedness would be the search for the sets of related tasks, and additionally the factors of variation that are useful. A useful formulation of this search then is in terms of finding the most useful factors of variation.

Given a set of tasks to learn to transfer over, the search for factors of variation can be carried out in many ways. These include statistical factors of variation, group theoretic disentanglement \citep{Pfau2020, Higgins2018}, or information theoretic \citep{Botvinick2015}. In the information theoretic case for example, the appropriate choice would be made in terms of the efficiency of coding. That is, there can be an assumption about the distribution from which tasks are sampled from, and the best notion of relatedness would be the most efficient way of encoding these tasks. 
\section{Conclusion}

In the present article, we introduced a framework that poses problems of learning to transfer from first principles. In particular, our framework explicitly exposes what it means for tasks to be related. We argued that relatedness requires a \emph{standard} against which to measure it, and learning to transfer amounted to finding a representation of the structure that defined such a standard. This structure can be written in terms of foliations. We then showed that modern problems, such as multitask learning, meta learning and transfer learning, can all be described within a coherent learning-to-transfer paradigm, and we specifically demonstrated how some example models implicitly make use of foliations. 

We believe that our framework will be crucial in studying and understanding problems of transfer using precise and interpretable techniques. In future, we will be investigating how this framework can applied to answering foundational questions regarding the transferrability between two arbitrary learning tasks, the complexity of learning to transfer, or  the development of novel methods of transfer.

\newpage
\acks{We would like to thank the School of Mathematics and Statisitics at the University of Sheffield for funding this project. We also thank So Takao, Laura Krasowska, Kate Highnam and Daniel Lengyal for the many discussions we had when developing and writing up this work.}
%  \newpage
\bibliography{main}
\newpage
\begin{appendices}
	
\section{Mathematical Preliminaries} \label{app:mathematical_prelims}

\subsection{Topology} \label{sec:topology}

%Why think about topologies?
We begin by describing what a topology is. In Section \ref{sec:space_of_learning_tasks} we argued that thinking about the topology of our set of related tasks is important. To understand the reasons for this, a rudimentary understanding of concepts in topology will be useful. 

%What is a topology
A topology is perhaps the simplest structure one can place on a set $\model$. 
\begin{definition} [Topology]
	Given a set $X$, a topology $\O_X$ on $X$ is a set of subsets of $X$ that satisfy the following conditions:
	\begin{enumerate}
		\item $X \in \O_X$ and $\emptyset \in \O_X$,
		\item $U_1, U_2 \in \O_X \implies (U_1 \cup U_2) \in \O_X$,
		\item $U_1, U_2, ..., U_n \in \O_X \implies \bigcap_{i=1}^n{U_i} \in \O_X$.
	\end{enumerate}
\end{definition}
Simply put, a topology $\Top_\model$ is a set of subsets of a set which is closed under intersections and unions, and also contains the null set $\emptyset$, and the set $\model$ itself. An element of a topology $U \in \Top_M$, which is a subset of $M$ is called an \emph{open set}, the difference of which with $M$ is called a \emph{closed set}. Despite these names, it is incorrect to think of an open set as an open interval; particularly, some sets can be both open and closed (for example, the full set $\model$). The tuple $(\model, \Top_\model)$ is called a topological space. 

\begin{figure}[ht] 
    \centering
    \includegraphics[width=0.45\textwidth]{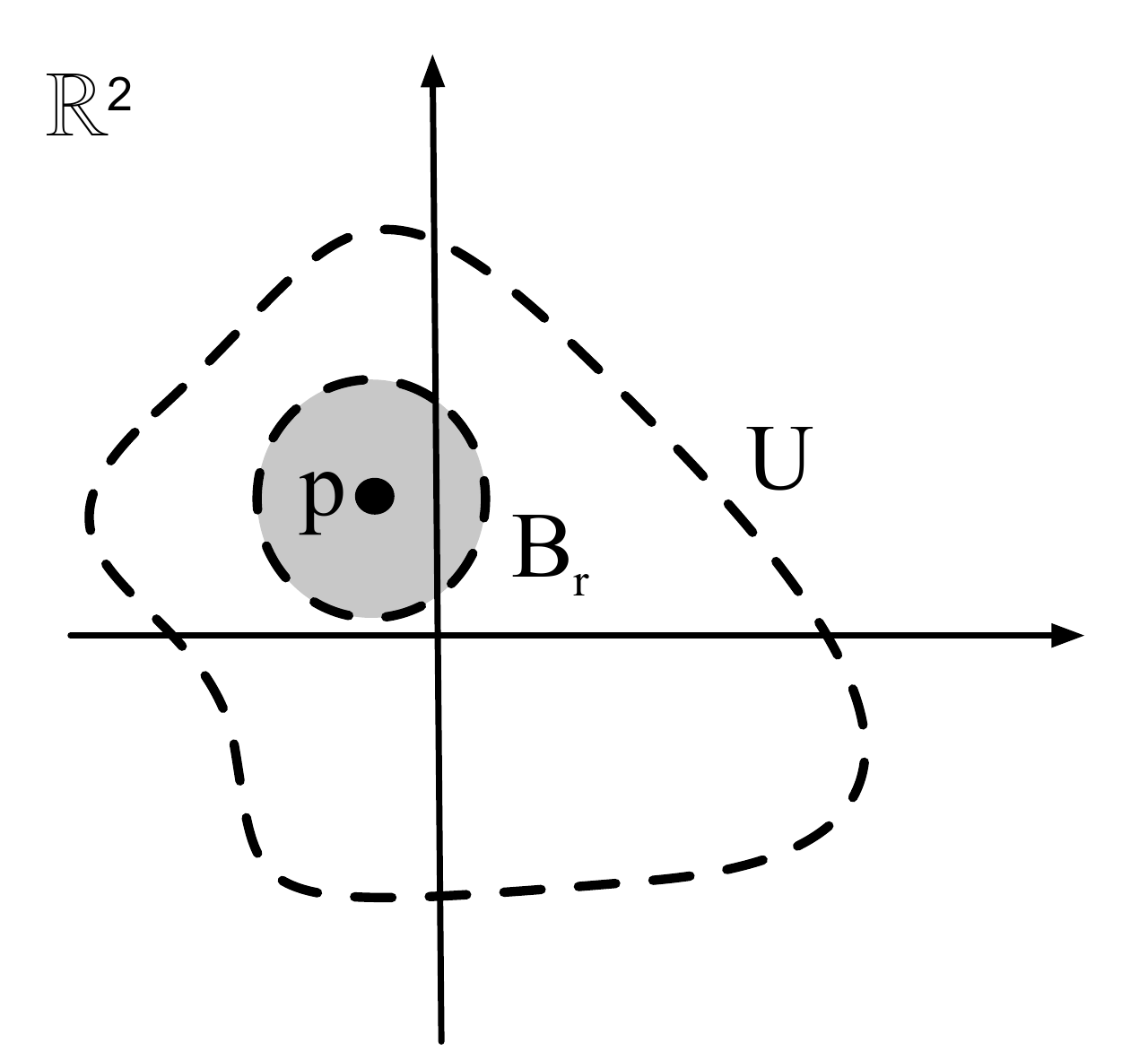}
    
    \caption{An open set in the standard topology on $\R^2$. Note that the dashed lines imply that the boundary is not included. Furthermore, the open set $U$ is populated with other open balls at all other points $p \in U$.}
	\label{fig:standard_topology}
\end{figure}

%Example of a topology
An intuitively familiar example of a topology is the called the standard topology $\Top^S_{\Rd}$ on $\Rd$. To define the standard topology, one defines its elements, the open sets implicitly as follows. An open set $U \in \Top^S_{\Rd}$ iff $\forall p \in U, \: \exists r \in [0, \infty] : \ball^r(p) \subseteq U$, where $\ball^r(p) := \big\{x \| \sqrt{\sum_{i=0}^d(x_i - p_i)^2}\big\}$ are open balls centred at $p$. This topology in $\R^2$ is depicted in Figure \ref{fig:standard_topology}.

%Definition of Neighbourhood
An important idea a topology affords us is a neighbourhood of a point. Different authors define this slightly differently, but these differences are inconsequential to our discourse. \citep{Lee2001} defines a neighbourhood of a point $p \in \model$ as an open set $U_p \in \Top_\model$ that contains $p$; on the other hand, \citep{Mendelson1990} define $A_p \subseteq \model$ to be a neighbourhood of $p$ if it completely contains an open set $U_p \in \Top_\model$ that contains $p$. Presently, we will stick to the former definition by \citep{Lee2001}, since it is simpler to consider.

%Neighbourhoods tell us about "closeness"
A neighbourhood of a point $p$ allows us to talk about which points are \emph{close} to $p$. That is, we can consider that all points in a neighbourhood $U_p$ are close to $p$ in this sense. Furthermore, consider 3 points $p, q, r \in \model$. Suppose that almost any neighbourhood $U_{pq}$ that contains $p$ and $q$ also contains $r$, but one can find many neighbourhoods $U_{pr}$ that contain $p$ and $r$, but do not contain $q$. We can consider this in terms of the size of $U_{pq}$ and $U_{pr}$ (using a measure), or say in terms of the radii of open balls that contain them. Then in this sense, we can say that $p$ and $r$ are qualitatively closer to each other than they are to $q$. 

%Example of a non-trivial topolgical space.
As an example, consider the circle space $\Circ^1$. Such an object can be defined by the space of equivalent classes of numbers on the real line, where the equivalence relation is given by $p \sim q$ if $p = q + 2\pi$. This is similar to taking the interval $[0, 2\pi]$ and gluing the 2 ends together. Topologically this means that $0.00001$ is \emph{close} to $2\pi - 0.00001$; this is of course different to real line, where these points are \emph{further away}. Thus, $\Circ^1$ is different to $\R$, and even the closed interval $[0, 2\pi]$. 

%Metric space vs topological space
At this point, it is worth noting the relationship between topological spaces and metric spaces. Loosely, a metric space is a set equipped with a metric, which is a numerical measure of distance between points; this is defined more formally in Definition \ref{def:metric}. 

\begin{definition}[Metric] \label{def:metric}
	Given a set $X$, a metric is a map $\rho: X \times X \rightarrow [0, \infty]$ that satisfies:
	\begin{enumerate}
		\item $\rho(x, y) = 0 \iff x = y$,
		\item $\rho(x, y) = \rho(y, x)$ (symmetry),
		\item $\rho(x, y) \leq \rho(x, z) + \rho(z, y)$ (triangle inequality),
	\end{enumerate}
	for $x, y, z \in X$.
\end{definition}

Every metric space defines a topological space through the metric topology; this is a topology similar to the standard topology, where the open balls are defined with respect to the metric on the metric space. In fact, the standard topology is the metric topology on $\R^d$ defined with respect to the Euclidean metric. There are, however topological spaces which do not have a corresponding metric; those that do, are called \emph{metrizable} \citep{Mendelson1990}. In this way, a topological space is more general. That is, as in the case of the circle example above, we are allowed to consider $closeness$ in terms of a metric, but it is not the only allowed way. Choosing a metric fixes the \emph{shape} of a space; for example, topologically, a geometric circle and a geometric ellipse are topologically equivalent to the circle space $\Circ^1$. 

%continuous functions are topology preserving maps.
Now, we can imagine transforming a circle into an ellipse by simply stretching or squashing actions (imagine a spherical balloon being squashed between 2 poles). Importantly, we do not tear (or create holes) the space. Such a transformation, that maps between 2 topological spaces is called a continuous map. More specifically, a continuous map is the topology preserving map; a map $f:X \rightarrow Y$ between 2 topological spaces $(X, \Top_X)$ and $(Y, \Top_Y)$ is \emph{continuous} if the pre-image of open sets in $Y$ is open in $X$. If such a map is invertible, $f$ is said to be a \emph{homeomorphism}, which is the special name given to topological isomorphisms. Thus, the map between a circle and an ellipse is a homeomorphism. 

%Continuous maps and \epsilon-delta
We note that in this way of defining a continuous map, the only necessary information about $X$ and $Y$ are their topologies. A typical way of defining a continuous map is in terms of the $\epsilon - \delta$ notation. Such a definition is only possible on a metric space, and we state that the definition given above is therefore more general. Of course, for a metrizable topological space, the $\epsilon - \delta$ definition follows once a suitable metric is found.

%Some important properties that topologies allow us to characterise
A topology allows us to then define a host of useful and important other notions that are preserved under continuous mappings. One such is the notion of compactness. Intuitively, one can think of a compact subset as a subset that is closed and bounded; for $\Rd$, the Heine-Borel Theorem \citep{HeineBorel1995} says that this is exactly true. A compact set remains compact under a continuous map. Another such property is that of (path)-connectedness. A subset $A$ is (path)-connected if one can draw a continuous line $\gamma_{pq}: [0, 1] \rightarrow A$ between any two points $p, q \in A$, where $\gamma(0) = p$ and $\gamma(1) = q$. Generally, path-connectedness and connectedness are distinct properties, but this distinction is not necessary in the present work, and we mean path-connectedness if we state that a set is connected.
\subsection{Manifolds}
A relatively simple, yet powerful object that we can start with is a topological manifold: 
\begin{definition} [Topological manifold] \label{def:topological_manifold}
A tuple $(\model, \Top_\model, \A_\model)$, is called a $d$-dimensional topological manifold if: 
\begin{enumerate}
	\item $\model$ is a set,
	\item $\Top_\model$ is a topology on $\model$, where $(U \in \Top_\model) \subseteq \model$ is called an open set. The topology must be Hausdorff and paracompact.
	\item $\A_\model$ is a collection of charts, where a chart $(U, \phi) \in \A_\model$ consists of an open set and a homeomorphism $\phi: U \rightarrow( U_{\Rd} \subseteq \Rd)$.
\end{enumerate}
Here, $\Rd$ is the usual real space endowed with $\Top^S_{\Rd}$, the standard topology.
\end{definition}

\begin{figure}[ht] 
    \centering
    \includegraphics[width=0.6\textwidth]{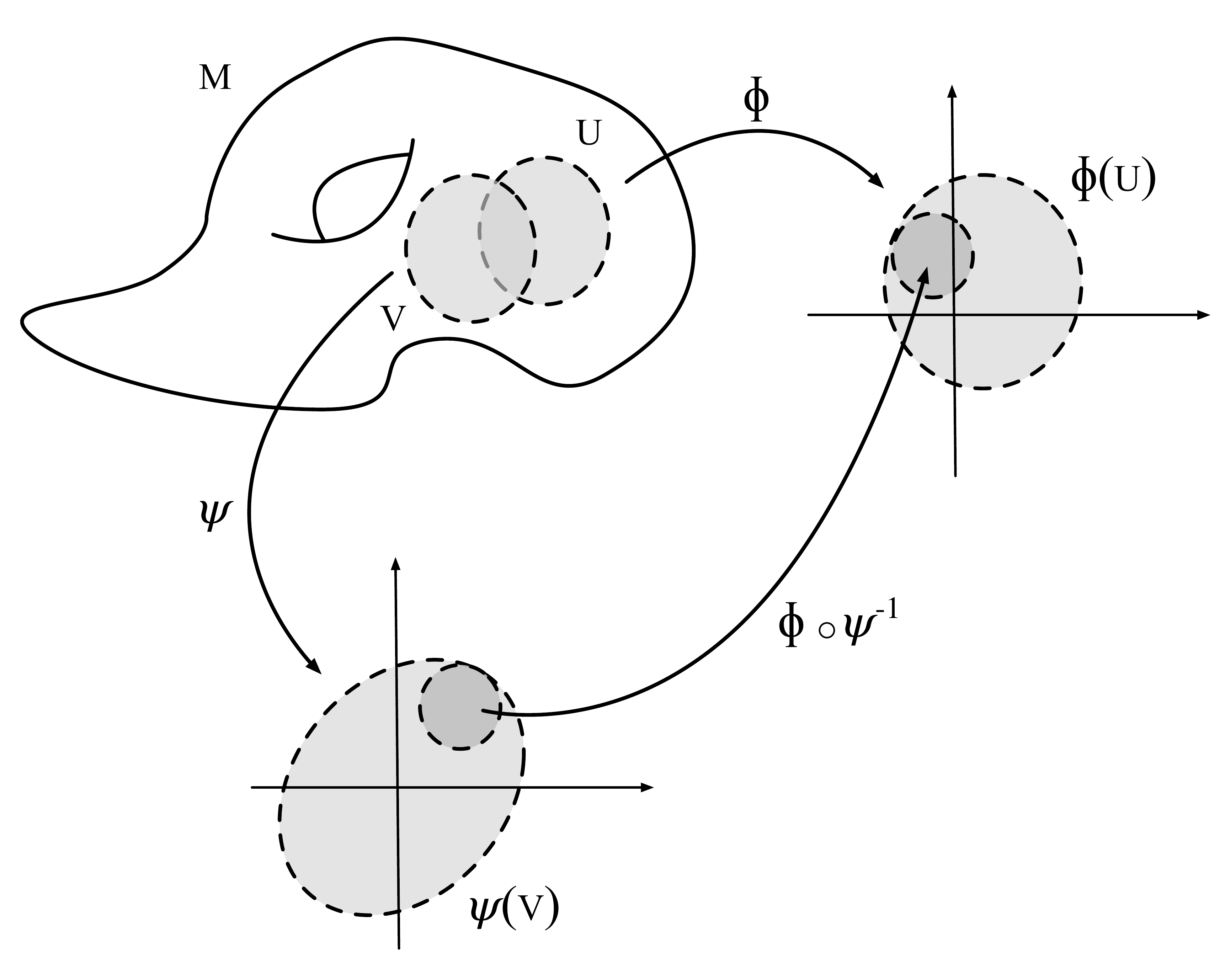}
    
    \caption{A topological manifold $\model$ with examples of its coordinate charts. We see here that despite being a complicated topological entity (see the hold in the middle), locally, it is equivalent to $\R^2$. Further, note that the chart transition map $\phi \circ \psi^{-1}: V_{\R^2} \rightarrow U_{\R^2}$ exists because a homeomorphism is invertible, and is continuous since the composition of continuous maps remains so.}
	\label{fig:topological_manifold}
\end{figure}

A topological manifold is created by endowing a topological space with a set of charts, called an atlas. A chart, from Definition \ref{def:topological_manifold} consists of a tuple of an open set $U \in \Top_\model$ and a map $\phi$ that is a homeomorphism between $U$ and a subset $U_{\Rd}$ of $\Rd$. This then is a generalisation of what we think of as the topological space $\Rd$; that is such a $d$ dimensional manifold is \emph{locally} homeomorphic to $\Rd$. The key aspect here is that the set $\model$ can be given any topology $\Top_\model$ on it, but we can still locally relate it to a space we are familiar with (the real space); see Figure \ref{fig:topological_manifold}. Each map $\phi$ that does this is called the \emph{coordinate map}. An atlas is a collection of such charts, such that the domains of each of the chart maps is a cover of $M$.

We note here that because open sets of the manifold are homeomorphic to $\Rd$, which is endowed with the standard topology, our manifold is locally compact. This means that every point on the manifold has a neighbourhood that is contained in a compact subset of X; intuitively, compactness gives us a notion of being closed (contains limit points) and bounded \citep{HeineBorel1995}. In addition to this, our manifolds are \emph{Hausdorff} \citep{Lee2001}. This means that if we take any 2 points on a manifold, there exists at least one neighbourhood of each point such that these subsets are non-intersecting. These properties give our manifolds nice properties that make them easy to work with. As an example, a trivial topology that can be given to a set $X$ is $\{X, \emptyset\}$. It is possible to show that every map onto $X$ is continuous, and as such this topology doesn't provide us with any useful structure; the Hausdorff property ensures that there are enough open sets to reason about interesting things.  

In Figure \ref{fig:topological_manifold}, there is a map that is defined to take elements in the intersection of two open sets from one chart to another. More precisely, given $U, V \in \Top_\model$ and charts on these sets $(U, \phi), (V, \psi) \in \A_\model$, where $\phi: U \rightarrow (U_{\Rd} \subset \Rd)$ and $\psi: V \rightarrow (V_{\Rd} \subset \Rd)$, and if $U \cap V \neq \emptyset$, then $\phi \circ \psi^{-1}: \psi(U \cap V) \rightarrow \phi(U \cap V)$. Such a map is called a chart transition map. These maps are important in defining differential properties on a manifold. 
\begin{definition}[$C^k$ - Atlas] \label{def:smooth_atlas}
An atlas $\A_\model$ of a topological manifold $\model$ is called a $C^k$-atlas if all chart transition maps are $k$ times continuously differentiable. 
\end{definition}
In this way, if our $\A_\model$ atlas satisfies these conditions, we call $\model$ a $C^k$-differentiable manifold. A smooth manifold then is a $C^\infty$-manifold; that is, it is infinitely differentiable. Such properties are defined to enable us to carry out calculus on manifolds in an invariant way; topologies don't provide enough restrictions because it is possible to construct maps that are homeomorphisms, but don't preserve differentiability. Furthermore, requiring that smoothness is preserved at the intersection of open sets also exposes another key aspect of the philosophy of manifolds, differential or otherwise. A manifold attempts to describe its properties in a coordinate independent way. That is, the manifold exists, with its properties, without the need for chart maps; the charts are our ways of interacting with the manifold. However, if the chart is valid, and respects the conditions we put on it, then our choice of a chart shouldn't change the properties of the underlying manifold. This is because if there are two charts in our atlas, $(U, \phi), (U, \psi) \in \A_\model$, which are based on the same open set $U \in \Top_\model$, then doing either chart transition $\phi \circ \psi^{-1}$ or $\psi \circ \phi^{-1}$ should not alter any properties we see on either coordinate. Thus, the condition in Definition \ref{def:smooth_atlas} ensures that smoothness properties are preserved between coordinates, and allows us to measure whether they can be preserved through maps between differential manifolds. A map $f$ on $M$ is said to be smooth, if $f \circ \phi^{-1}$ is smooth, for any $(U, \phi) \in \A_\model$. A map that does preserve smoothness, and is invertible is called a \emph{diffeomorphism}. In this way, the chart transition maps are therefore perhaps the most obvious examples of diffeomorphisms. 

As a slight tangent, this philosophy contributes to the beauty of differential geometry. It allows us to define properties intrinsic to a space without worrying about how we choose to represent them; these properties are then independent of the coordinate system we choose. This is of course a very desirable attribute, since we don't want to worry about whether our theories are the way they are because of a particular choice in coordinates that we made.

\subsection{Groups and Group Actions} \label{app:groups}
\begin{definition}[Group]
	A set $\G$ with an operation $\bullet: \G \times \G \rightarrow \G$ is called a group $(\G,\bullet)$  if:
	\begin{enumerate}[label=\alph*)]
		\item $\bullet$ is associative
		\item there exists $e \in \G$ such that $e \bullet g = g$ for any $g \in \G$, and
		\item $\forall g \in \G$ there exists $g^{-1} \in \G$ such that $g \bullet g^{-1} = e$ and $g^{-1} \bullet g = e$.
	\end{enumerate}
\end{definition}
A group is said to be commutative or \emph{Abelian} if $g \bullet h = h \bullet g$. 

The key way in which we will be using a group is through its action on another set. 
\begin{definition}[Group action (left)]
	The (left) action of a group $(\group,\bullet)$ on a set $\X$ is a map $\rho: \group \times X \rightarrow X$ that satisfies:
	\begin{enumerate}[label=\alph*)]
		\item $\rho(h, \rho(g, x)) = \rho(h \circ g, x)$,
		\item $\rho(e, x) = x$,
	\end{enumerate} 
	for $g, h \in \group$, $e \in \group$ is the neutral element in $\group$, and $x \in X$.
\end{definition}
A right action reverses the order of the input; we brevity, we will only consider left group actions, and therefore won't qualify which we use. A group action gives us a way to traverse a set in terms of a set of transformations. That is, we can think of a group as a set of transformations, and the action of an element of a group is applying a particular transformation an element of another group. An example of a group is SO(3), which is group of all spherical rotations; an element of SO(3) can act on an element of $\R^3$ by rotating it, in the intuitive sense. 

Groups play an important role in the idea of symmetries and invariances \citep{Weyl2015}. A particular reason for that are the two conditions that a group action must satisfy. These conditions ensure a sort of consistency in the behaviour of the action. Consider the following scenario. We have three points $x, y, z \in X$. Suppose $y = g_{xy}(x)$, $z = g_{yx}(y)$ and the $z = g_{xz}(x)$. The first condition ensures that $g_{xz} = g_{xy}\bullet g_{yz}$, that we can blindly choose from $\G$ whether to move from $x$ to $z$ directly, or to do so through an intermediate point $y$. The second condition ensures this consistency holds for the neutral element of the group too. 

% Write something about groups and equivalence relations
% A group, together with its group action defines a partitioning of a set $X$.
%
\begin{definition}[Orbit]
	Given a group $\G$ and its action $\rho$ on a set $X$, the orbit of $\G$ centered at a point $x \in X$, which is denoted by $O_\group(x)$, is a set defined as:
	\begin{equation*}
		O_\group(x) = \big\{y \| y = \rho(g, x) \: \forall g \in \group \big\}.
	\end{equation*}
\end{definition}
Since $\G$ includes the neutral element, $x \in O_\G(x)$. The consistency requirements of a group action, together with the requirements of a group imply that an orbit forms a well defined equivalence class. That is, if $x, y \in O_\G(x)$, then $O_\G(x) = O_\G(y)$. In other words, there is an equivalence relation, where $x \sim y$ if $\exists g \in \G$ such that $x = g(y)_\G$, which is of course used in the definition of an orbit. With an equivalence relation comes a quotient space $X/\sim$; this is the space of equivalence classes. The quotient space defined by orbits of a $\G$ is denoted by $X/\G$ and is called the \emph{orbit space}. 

A group action is called \emph{free} if its isotropy groups are trivial, and only contain the neural element. This implies that for a free group action, if $x, y \in O_\G(x)$, then $O_\G(x) = O_\G(y)$. This means that the orbits of a free action are well defined equivalence classes. 

An important class of groups that we consider are Lie Groups \citep{Nakahara2003, Lee2001,Hall2015}. Lie groups are simply groups that are also smooth manifolds. 

\subsection{Pseudogroups} \label{app:pseudogroups}
A pseudogroup is a generalisation of a group, often written for local diffeomorphisms, where appropriate. They play an important role in partial symmetries \citep{Lawson1998}.
\begin{definition}[Pseudogroups] \label{def:pseudogroup} \citep{Lisle1998} 
	If $M$ is a smooth manifold, and $\G$ is a collection of local diffeomorphisms on open subsets of $M$ into $M$, then $\G$ is a pseudogroup if:
	\begin{enumerate}[label=\alph*)]
		\item $\G$ is closed under restrictions: if $\tau: U \rightarrow M$, for $U$ an open set of $M$, is in $\G$, then $\tau|_V \in \G$ for any $V \subseteq U$ is also in $\G$,
		\item if $U \subseteq M$ is an open set, and $U = \bigcup_sU_s$, and $\tau: U \rightarrow M$, with $\tau|_{U_s} \in \G$, then $\tau \in \G$,
		\item $\G$ is closed under composition: if $\tau_1, \tau_2 \in \G$, then $\tau_1 \circ \tau_2 \in \G$, whenever the composition is defined (i.e over intersections of the appropriate domain and codomain),
		\item $\G$ contains an identity diffeomorphism over $M$,
		\item $\G$ is closed under inverse: if $\tau \in \G$, then $\tau^{-1} \in \G$, with a domain $\tau(U)$, where $U$ is the domain of $\tau$.
	\end{enumerate}
\end{definition}

We can see that the latter 3 conditions look similar to what we would expect from a group of differmorphisms. A pseudogroup allows for such structure when the diffeomorphisms aren't defined globally.
%\paragraph{Sets and Topologies}
%\paragraph{Manifolds: Topological and Differential}
%\paragraph{Foliations}
%\paragraph{Example of foliation: Spacetime}
%
%\subsection{An intuitive explanation of why foliations make sense}
%\begin{itemize}
%	\item Start with why we can think of the parameter space of models; these are representations of the `things' on the manifold. These representations or models then give us the natural topology. 
%	\item Then we can introduce foliations as a way of adding `structure' to this manifold of representations. 
%	\item Foliations allow for a way to define a logical separation/hierarchy of `things'. It can allow us to mathematically talk about what a class of things is, what is a variation of that thing, what thing is not a member of that class, and also topological and geometric properties of the classes. 
%	\item Can allow us to make statements like `black is the opposite of white in terms of color'. They are also opposites in terms of `mood'. The foliations are made to satisfy these conditions of separation due to RBB, or due to mood. 
%\end{itemize}

\section{Proofs}
\subsection{Induced topology on the Task space} \label{sec:topology_proof}
\begin{proof}[Theorem \ref{theo:topology}]
	To complete this proof, we need to show the following: 
	
	\begin{enumerate}
		\item $\task \in \O_\task$ and $\emptyset \in \O_\task$: True by definition.
		\item $(U_1 \cup U_2) \in \O_\task$ for $U_1, U_2 \in \O_\task$: For a $t \in U_1 \cup U_2$, there exists $\mathcal{B}_{t,1}(\epsilon_1) \subseteq U_1$ or $\mathcal{B}_{t,2}(\epsilon_2) \subseteq U_2$ by construction of $U_1$ and $U_2$. Thus, the condition is satisfied.
		\item  $\bigcap_{i=1}^n{U_i} \in \O_\task$ for $U_i \in \O_\task$: For a $t \in U_1 \cap U_2$, there exists $\mathcal{B}_{t}(\epsilon_i) \subseteq U_i$ by construction of $U_i$. Suppose $\epsilon^* = \min(\epsilon_1, \epsilon_2, ..., \epsilon_n)$. Then, $\mathcal{B}_t(\epsilon^*) \subseteq  \bigcap_{i=1}^n{U_i}$. This follows from the assumption of continuity of the loss function w.r.t. to the model space and $[0, \infty]$.
	\end{enumerate}
\end{proof}

\subsection{Relatedness from $\foliation$} \label{sec:relatedness_foliation}
\begin{proof}[Theorem \ref{theo:relatedness_foliation}]
	A leaf $\leaf$ is an immersed, connected, smooth submanifold of a manifold on which $\foliation$ is defined \citep{Lee2001,Stefan1974}. Thus, we need to show that, on such $\leaf$, there exists a countable (pseudo)group of transformations. Assume that $\leaf$ has a dimension of $d$.

	To begin with, we will identify some properties of the topology on $\leaf$. Assume that $\leaf$ has subset topology derived from the foliation \citep{Camacho2013}. From \citet[Theorem~1.10]{Lee2001}, we know that $\leaf$ has a countable topological basis $\ball$ of coordinate balls. Take the domains of the charts of $\foliation$ restricted to $\leaf$ (denoted by $\foliation|_\leaf$); these are open subsets of $\leaf$, and form an open cover. We can create a refinement of this cover, where for each domain $U \in \foliation|_\leaf$, we find $B_U \subseteq \ball$ where $\bigcup_{\ball^i \in B_U}\ball^i = U$. By restricting the chart maps to the appropriate $\ball^i$, we now have an altas on $\leaf$ that is made up of coordinate balls. We will denote this as $\atlas_\ball = \{(\ball^i, \phi_i)\}$, where $\phi_i$ is the restriction of the appropriate chart map; we have simplified the notation here by ignoring its construction from the original chart induced by the foliation.  Further, if we write $\ball^i \in \atlas_\ball$, we mean the domain of a chart in $\atlas_\ball$. 

	We also know that $\leaf$ is connected. Thus, given any two points on $p, q \in \leaf$, we can find a sequence $\ball_1, ..., \ball_n \in \ball$ such that $p \in \ball_1$ and $q \in \ball_n$, and $\ball^i \cap \ball_{i-1} \neq \emptyset$. One can think of this as a sequence of connected coordinate balls that connect $p$ and $q$. 

	Next, let us look at the Euclidean space $\Rd$, equipped with the standard topology. Let us equip $\Rd$ with the standard frame. That is, the standard frame $V = (V_1, ..., V_d)$ is a collection of vector fields $V_i$, where the $j$-th component of the vector $V_i(x)$ is given by $V^j_i(x) = \delta^j_i$ for $x \in \Rd$. Note that since $V$ is a global frame, we have that $(V_1(x), ..., V_d(x))$ is a basis for the tangent space at $x \in \Rd$. 

	We know that $V$ is a global frame, and that each $V_i$ is a complete vector field. Thus the flow $\theta_i: \R \times \Rd \rightarrow \Rd$ exists. Then, $\theta_i(t): \Rd \rightarrow \Rd$, for some $t \in \R$ is a diffeomorphism. We define the following map $\tau_{(x_1, ..., x_d)}: \Rd \rightarrow \Rd$ as,
	\begin{equation} \label{eqn:rd_transformation}
		\tau_{(x_1, ..., x_d)}(x) = \theta_d(x_n) \circ ... \circ \theta_1(x_1)(x).
	\end{equation}
	where $x_i \in \R$. For all $(x_1, ..., x_d) \in \Rd$, the transformations $\tau_{(x_1, ..., x_d)}$ form an additive group. It's inverse is given by,
	\begin{equation}
		\tau_{(x_1, ..., x_d)}^{-1}(x) = \tau_{(-x_1, ..., -x_d)}(x) = \theta_d(-x_n) \circ ... \circ \theta_1(-x_1)(x).
	\end{equation}
	Furthermore, this group is transitive in $\Rd$, since we can find a transformation from any point $x \in \Rd$ to any other point $y \in \Rd$.

	Suppose $x = (x_1, ..., x_d)$ and $y = (y_1, ..., y_d$). The following is then known to be true. 
	\begin{equation}
		y = \theta_n(y_n - x_n) \circ ... \circ \theta_1(y_1-x_1)(x),
	\end{equation}
	Let us denote this as a transformation $\tau_{xy}$. Note that there is an inverse $\tau_{xy}^{-1} = \tau_{yx} = \theta_n(x_n - y_n) \circ ... \circ \theta_1(x_1-y_1)$, where,
	\begin{equation}
		x = \theta_n(x_n - y_n) \circ ... \circ \theta_1(x_1-y_1)(y).
	\end{equation}
	Thus, given the coordinates of any two points, we can construct a diffeomorphism between them using the standard frame. 

	Let us go back to our leaf $\leaf$. Each $\ball^i \in \atlas_\ball$ is homeomorphic to a standard ball in $\Rd$. That is for each $\ball^i$, there is a $\ball_{\Rd}^r$, for some $r \in \R^+$, that it is homeomorphic to, and this homeomorphism is given by the chart map. Each such coordinate ball is itself homeomorphic to $\Rd$ via a map such as,

	\begin{equation} \label{eqn:homeomorphism}
		h_i(x_i) = \frac{x_i}{r - ||x||} = y_i,
	\end{equation}
	for an $x \in \ball_{\Rd}^r$ and $y \in \Rd$. Note that $h_i$ exists for each of the coordinates. The inverse of this map is given by,
	\begin{equation} \label{eqn:homeomorphism_inverse}
		h_i^{-1}(y_i) = \frac{ry_i}{1 + ||y||} = x_i.
	\end{equation}
	Note that these maps are smooth. Then, we can construct a diffeomorphism $\pi_{(x_1, ..., x_d)}^i: \ball^i \rightarrow \ball^i$, for $x_1, ..., x_d \in \R$  as,
	\begin{equation} \label{eqn:ball_transformation}
		\pi_{(x_1, ..., x_d)}^i(p) = \phi_i^{-1} \circ h_i^{-1} \circ \tau_{(x_1, ..., x_d)} \circ h_i \circ \phi_i(p).
	\end{equation}
	The inverse of this map is,
	\begin{equation} \label{eqn:ball_transformation_inverse}
		{\pi_{(x_1, ..., x_d)}^{-1}}^{i}(p) = \phi_i^{-1} \circ h_i^{-1} \circ \tau_{(x_1, ..., x_d)}^{-1} \circ h_i \circ \phi_i(p).
	\end{equation}
	Then, for any $p, q \in \ball^i$, we can construct a map $\pi^i_{pq}$,
	\begin{equation} \label{eqn:ballpoint_transformation}
		\pi^i_{pq}(p) = \phi_i^{-1} \circ h_i^{-1} \circ \tau_{xy} \circ h_i \circ \phi_i(p) = q,
	\end{equation}
	where $x = h_i \circ \phi_i(p)$ and $y = h_i \circ \phi_i(q)$. The inverse ${\pi^{-1}_{pq}}^{i} = \pi^i_{qp}$ of this map from $q \in \ball^i$ to $p \in \ball^i$ is given by,
	\begin{equation} \label{eqn:ballpoint_transformation_inverse}
		\pi^i_{qp}(q) = \phi_i^{-1} \circ h_i^{-1} \circ \tau_{yx} \circ h_i \circ \phi_i(p) = p.
	\end{equation}
	$\pi_{(x_1, ..., x_d)}^i$ is then a group of diffeomorphisms that acts transitively on $\ball^i$. We have effectively used the homeomorphism defined in Equation \ref{eqn:homeomorphism} to move the transformations on $\Rd$ given in Equation \ref{eqn:rd_transformation} to transformation in $\ball^i$. Let us denote by $\Pi^i$ the set of all transformations of the form $\pi_{(x_1, ..., x_d)}^i$.
	
	An open subset of $\ball^i$ is denoted as $U^i$. We will also denote a countable (possibly finite) sequence of elements from $\Rd$ as $\sequence_{\Rd} = \{x^1, x^2, ...\}$, and a countable sequence of coordinate balls $\sequence_{U^i, \ball} = \{U^i, \ball_2, ... \}$, where $\ball_j \in \atlas_\ball$. Now, we define a set of transformations $\Pi$, where each $\pi \in \Pi$ is a map $\pi^{\sequence_{U^i, \ball}}_{\sequence_\Rd}: U^i \rightarrow \leaf$, given by,
	\begin{equation} 
		\pi^{\sequence_{U^i, \ball}}_{\sequence_\Rd} = ... \circ \pi^{\ball_j}_{x_j} \circ ... \circ \pi^{\ball_2}_{x_2} \circ \pi^{U_i}_{x_1}(p),
	\end{equation}
	such that the image and domain of each subsequent transformation is valid. That is, the intersection of image and domain $\mathrm{im}(\pi^{\ball_j}_{x_j}) \cap \mathrm{dom}(\pi^{\ball_{j+1}}_{x_{j+1}}) \neq \emptyset$. Thus $\Pi$ contains all transformations acting on the leaf $\leaf$ that can be written as countable compositions of maps from any $\Pi^i$, as long as the ranges and domains make sense. 
	
	We now show that $\Pi$ satisfies each condition in Definition \ref{def:pseudogroup}.
	\begin{enumerate}[label=\alph*)]
		\item Satisfied by definition, since $\mathrm{dom}(\pi^i) = \ball_q$, $\pi^i|_U$ for $U \subset \ball^i$ exists,
		\item follows from (1),
		\item since all compositions are included, this follows from definition,
		\item by settings $x_i = (0, ..., 0)$ for all $x_i$ in $\sequence_\Rd$, we obtain the identity maps, 
		\item follows from Equation \ref{eqn:ball_transformation}.
	\end{enumerate}
	
	Finally, we must show that $\Pi$ is transitive. We do this be showing constructively that between any two points on $\leaf$, we can find a map between them in $\Pi$. Recall that due to the connectedness of $\leaf$, we can find a sequence $\{\ball_n\}_{pq}$ of coordinate balls that connect two points $p, q \in \ball^i$. Take the intersection $\ball^i \cap \ball^i-1$. We will denote by $p_{i, i-1}$ to be any point in this intersection. Then we can define a transformation $\pi_{pq}$ between points $p, q \in \leaf$ as,
	\begin{equation} 
		\pi_{p, ..., p_{i, i-1}, ...,  q}(p) = \pi_{p_{n, n-1},q} \circ ... \circ \pi_{p, p_{2, 1}}(p) = q.
	\end{equation}
	It's inverse is given by,
	\begin{equation} 
		\pi_{q, ..., p_{i, i-1}, ...,  p}(q) = \pi_{p_{2, 1},p} \circ ... \circ \pi_{q, p_{n, n-1}}(q) = p.
	\end{equation}
\end{proof}
\subsection{MAML} \label{sec:maml_proof}
\begin{proof}[Theorem \ref{theo:maml_simple}]
	Since the loss function is given by $L((t_1, t_2), (m_1, m_2)) = \sum_{i = 1}^2(m_i - t_i)^2$, assuming that the coordinate system is centered at $m_0$, the vector field on $\model$ is given by: 
	\begin{equation}
		\mathrm{M}((m_1, m_2)) = (\dot{m_1}, \dot{m_2})= 2m_1\frac{\partial}{\partial m_1} + 2m_2\frac{\partial}{\partial m_2}.
	\end{equation}
	This is a system of differential equations which can be solved to give us a flow $\theta(k) = (m_1(k), m_2(k))$; the initial value condition is $m_i(0) = 0$, since we have centered the coordinate system at $m_0 = (0, 0)$.

	The flow is given by:
	\begin{equation} \label{eq:maml_orbit}
		m_i(k) = t_ie^{-2tk} + t_i.
	\end{equation}

	Since $L(t, m_t^*) = 0$, we have that for the $m(k)$ at which $\|L(t, m(k)) - L(t, m_t^*)\| = \epsilon$, $L(t, m(k)) = \epsilon$.
	Thus, 
	\begin{equation} \label{eq:maml_proof_0}
		\sum_{i=1}^2((t_ie^{-2tk} + t_i) - t_i)^2 = \epsilon.
	\end{equation}
	This can be rearranged to produce
	\begin{equation}
        \sum_{i=1}^2t_i^2 = \frac{\epsilon}{e^{-4k}}. 
    \end{equation}
\end{proof}
\begin{proof}[Corollary \ref{coro:maml_simple}]
	Equation \ref{eq:maml_proof_0} can be solved for a $k$, which gives us: 
	\begin{equation}
        k_\epsilon = 4\ln\bigg(\frac{\epsilon}{t_1^2 + t_2^2}\bigg).
    \end{equation}
	This can then be plugged into Equation \ref{eq:maml_orbit} to give us:
	\begin{equation}
        m_i = -t_i\bigg(\frac{\epsilon}{t_1^2 + t_2^2}\bigg) + t_i.
    \end{equation}
\end{proof}
\end{appendices}
\end{document}